\documentclass{article}

\usepackage{microtype}
\usepackage{graphicx}
\usepackage{subcaption}
\usepackage{booktabs} 

\usepackage{hyperref}

\usepackage{tikz}
\usetikzlibrary{arrows.meta,positioning,fit,decorations.pathreplacing}

\usepackage{multirow}

\usepackage[preprint]{icml2026}

\usepackage{amsmath}
\usepackage{amssymb}
\usepackage{mathtools}
\usepackage{amsthm}

\usepackage[capitalize,noabbrev]{cleveref}

\theoremstyle{plain}

\theoremstyle{definition}

\theoremstyle{remark}

\usepackage[disable,textsize=tiny]{todonotes}

\icmltitlerunning{Contrastive Concept-Tree Search}

\begin{document}

\twocolumn[
  \icmltitle{Contrastive Concept-Tree Search for LLM-Assisted Algorithm Discovery}



  \icmlsetsymbol{equal}{*}

  \begin{icmlauthorlist}
    \icmlauthor{Timothee Leleu}{equal,ntt,sta}
    \icmlauthor{Sudeera Gunathilaka}{ais}
    \icmlauthor{Federico Ghimenti}{sta}
    \icmlauthor{Surya Ganguli}{sta}
  \end{icmlauthorlist}

  \icmlaffiliation{ntt}{NTT Research, Sunnyvale, CA, USA}
  \icmlaffiliation{sta}{Stanford University, Palo Alto, USA}
  \icmlaffiliation{ais}{AIST, Tsukuba, Japan}

  \icmlcorrespondingauthor{Timothee Leleu}{timothee.leleu@ntt-research.com,tleleu@stanford.edu}

  \icmlkeywords{LLM-assisted algorithm discovery, black-box optimization, evolutionary algorithm, contrastive learning, Tree-structured Parzen Estimator, interpretability}

  \vskip 0.3in
]



\printAffiliationsAndNotice{}  

\begin{abstract}
Large language Model (LLM)-assisted algorithm discovery is an iterative, black-box optimization process over programs to approximatively solve a target task, where an LLM proposes candidate programs and an external evaluator provides task feedback. Despite intense recent research on the topic and promising results, how can the LLM internal representation of the space of possible programs be maximally exploited to improve performance is an open question. Here, we introduce Contrastive Concept-Tree Search (CCTS), which extracts a hierarchical concept representation from the generated programs and learns a contrastive concept model that guides parent selection. By reweighting parents using a likelihood-ratio score between high- and low-performing solutions, CCTS biases search toward useful concept combinations and away from misleading ones, providing guidance through an explicit concept hierarchy rather than the algorithm lineage constructed by the LLM. We show that CCTS improves search efficiency over fitness-based baselines and produces interpretable, task-specific concept trees across a benchmark of open Erdős-type combinatorics problems. Our analysis indicates that the gains are driven largely by learning which concepts to avoid. We further validate these findings in a controlled synthetic algorithm-discovery environment, which reproduces qualitatively the search dynamics observed with the LLMs.
\end{abstract}

\section{Introduction}

Recent advances in large language models (LLMs) have enabled a new class of algorithm-discovery systems in which program synthesis is embedded in an iterative black-box optimization loop: candidate programs are proposed by an LLM, evaluated by an external objective function, and refined over successive rounds. Such an iterative refinement is often framed as an evolutionary process over the space of possible programs, with the LLM acting as a mutation operator and the programs' performance playing the role of a fitness function~\cite{Lehman2024}. Recent applications of this paradigm have led to the rediscovery or improvement of solutions to Erd\H{o}s-style problems in combinatorics and related areas \cite{novikov2025alphaevolve,georgiev2025mathematical}, and have produced heuristics that compete with highly engineered human designs, including strong SAT-solver variants \cite{yu2025autonomous}. Overall, these results suggests that LLM-assisted discovery can generate novel, nontrivial algorithms, but at the same time, assessing the originality and impact of the solutions found is inherently challenging: validation often requires substantial domain expertise and collaborative effort, and it remains unclear to what extent current successes reflect general algorithmic advances rather than problem-specific improvements!\cite{ai-erdos-2026}. Nevertheless, it seems unlikely that current approaches are maximally exploiting the structure of the  LLM's representation of the program space, within which the automated search takes place. Precisely because the empirical picture is still evolving, this is a timely moment to examine how LLM-assisted algorithm-discovery might be structured to scale more reliably, and to understand which search mechanisms are likely essential as these methods are applied to harder and less well-understood problems.

Most existing LLM-assisted algorithm-discovery systems rely on a fitness-driven evolutionary updates: parents are selected based on their fitness on the target task and the iterative search proceeds as a walk over the lineage of programs ordered by performance. While effective in practice, this strategy operates directly within the program space, which remains weakly structured. Even when constrained by an LLM, the algorithms produced do not admit a clear notion of locality, smooth variation, or compositional structure that can be exploited by the search, limiting optimization to generic black-box heuristics. As a result, guidance comes largely from raw fitness signals and the LLM’s implicit prior, rather than from an explicit, learned understanding of which semantic components of the algorithms are actually useful. This is a key limitation for genuinely novel problems, where prior knowledge may be insufficient and usefulness must instead be discovered by interacting with the task.

The key idea of this work is to make the latent structure of the algorithm space explicit. We posit that algorithms can be described in terms of an underlying semantic concept space organized as a hierarchy, where each node represents a concept and the edges encode refinement or specialization. Each concept contributes positively or negatively to task performance, and strong algorithms arise from combining useful concepts while avoiding harmful ones. Such a tree-structured organization is natural in scientific discovery, which is mediated by shared concept hierarchies (e.g., the Mathematics Subject Classification or the American Physical Society taxonomy).

Building on this hypothesis, we guide search using contrastive statistics that identify which concepts are associated with high-performing algorithms. Instead of navigating program space directly, we bias parent selection and semantic edits toward empirically useful regions of the concept tree, enabling a more structured and data-driven search.

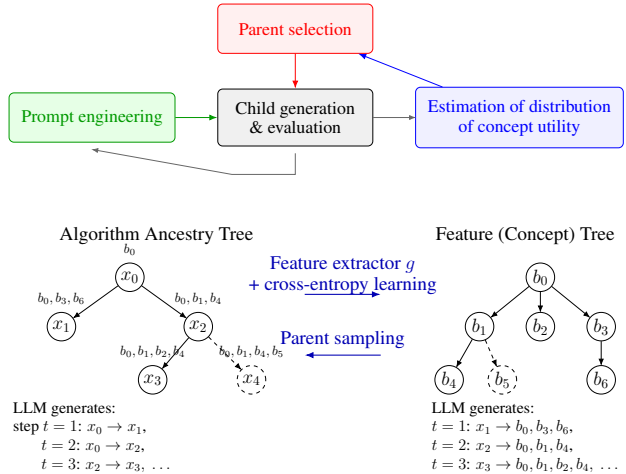
\begin{figure}[h]
\centering
\resizebox{\columnwidth}{!}{%

\begin{tikzpicture}[
  font=\Large,
  >=Latex,
  node distance=10mm and 12mm,
  big/.style={
    draw,
    rounded corners,
    align=center,
    inner sep=10pt,
    minimum width=44mm,
    minimum height=14mm
  },
  sel/.style={big, draw=red,  text=red,  fill=red!6},
  child/.style={big, draw=black, text=black, fill=black!6},
  prompt/.style={big, draw=green!60!black, text=green!60!black, fill=green!10},
  infer/.style={big, draw=blue, text=blue, fill=blue!6},
  arrSel/.style={->, draw=red, thick},
  arrPrompt/.style={->, draw=green!60!black, thick},
  arrInfer/.style={->, draw=blue, thick},
  arrFB/.style={->, draw=black!60}
]

\node[sel]   (Bsel)   {Parent selection};
\node[child, below=of Bsel] (Bchild) {Child generation\\\& evaluation};
\node[prompt, left=of Bchild] (Bprompt) {Prompt engineering};
\node[infer, right=of Bchild] (Binfer) {Estimation of distribution\\of concept utility};

\draw[arrSel]    (Bsel) -- (Bchild);
\draw[arrPrompt] (Bprompt) -- (Bchild);
\draw[arrInfer]  (Binfer) -- (Bsel);

\draw[arrFB]     (Bchild) -- (Binfer);

\draw[arrFB]
  ([yshift=-2mm]Bchild.south)
  -- ++(0,-6mm)
  -- ++(-18mm,0)
  -- ([yshift=-2mm]Bprompt.south);

\end{tikzpicture}
}\par\vspace{6mm}\par
\resizebox{\columnwidth}{!}{%


\begin{tikzpicture}[
  font=\Large,
  >=Latex,
  node distance=14mm and 18mm,
  circ/.style={circle,draw,minimum size=7.5mm,inner sep=0pt},
  dashedcirc/.style={circ,dashed},
  edgelab/.style={midway,fill=white,inner sep=1.2pt,font=\Large},
  blueTxt/.style={text=blue!70!black,font=\Large},
  arrowTxt/.style={blueTxt,align=center}
]

\node[font=\Large] at (-4.9,3.1) {Algorithm Ancestry Tree};

\node[circ] (x0) at (-5.6,2.0) {$x_0$};
\node[circ] (x1) at (-7.4,0.7) {$x_1$};
\node[circ] (x2) at (-3.8,0.7) {$x_2$};
\node[circ] (x3) at (-5.0,-0.7) {$x_3$};
\node[dashedcirc] (x4) at (-2.4,-0.7) {$x_4$};

\node[font=\normalsize, above=2pt of x0] {$b_0$};
\node[font=\normalsize, above=2pt of x1] {$b_0,b_3,b_6$};
\node[font=\normalsize, above=2pt of x2] {$b_0,b_1,b_4$};
\node[font=\normalsize, above=2pt of x3] {$b_0,b_1,b_2,b_4$};
\node[font=\normalsize, above=2pt of x4] {$b_0,b_1,b_4,b_5$};

\draw[->] (x0) -- (x1);
\draw[->] (x0) -- (x2);

\draw[->] (x2) -- (x3);
\draw[->,dashed] (x2) -- (x4);

\node[align=left,font=\large] at (-6.6,-2.2) {LLM generates:\\
step $t=1$: $x_0 \rightarrow x_1$,\\
\hspace*{7.4mm}$t=2$: $x_0 \rightarrow x_2$,\\
\hspace*{7.4mm}$t=3$: $x_2 \rightarrow x_3,\ \ldots$};

\node[arrowTxt] at (0.0,2.05)
  {Feature extractor $g$\\+ cross-entropy learning};
\draw[->,blue!70!black,thick] (-1.0,1.55) -- (1.0,1.55);

\node[arrowTxt] at (0.0,0.35) {Parent sampling};
\draw[->,blue!70!black,thick] (1.0,-0.05) -- (-1.0,-0.05);

\node[font=\Large] at (4.8,3.1) {Feature (Concept) Tree};

\node[circ] (b0) at (5.2,2.0) {$b_0$};
\node[circ] (b1) at (3.6,0.7) {$b_1$};
\node[circ] (b2) at (5.2,0.7) {$b_2$};
\node[circ] (b3) at (6.8,0.7) {$b_3$};

\node[circ] (b4) at (2.8,-0.7) {$b_4$};
\node[dashedcirc] (b5) at (4.2,-0.7) {$b_5$};
\node[circ] (b6) at (6.8,-0.7) {$b_6$};

\draw[->] (b0) -- (b1);
\draw[->] (b0) -- (b2);
\draw[->] (b0) -- (b3);

\draw[->] (b1) -- (b4);
\draw[->,dashed] (b1) -- (b5);
\draw[->] (b3) -- (b6);

\node[align=left,font=\large] at (4.8,-2.2) {LLM generates:\\
$t=1$: $x_1 \rightarrow b_0,b_3,b_6$,\\
$t=2$: $x_2 \rightarrow b_0,b_1,b_4$,\\
$t=3$: $x_3 \rightarrow b_0,b_1,b_2,b_4,\ \ldots$};

\end{tikzpicture}

}
\caption{(top) Overview of our algorithm-discovery loop: given a task and evaluator, we repeatedly sample a parent program from an archive, prompt the LLM to generate a mutated child, and evaluate it. The child’s outcome updates both the prompt context and a growing tree of semantic concepts, which then biases parent selection in subsequent iterations. (bottom) Repeating this process yields a phylogenetic program lineage $(x_0,x_1,\ldots)$ and an induced concept tree $(b_0,b_1,\ldots)$. We fit concept utility models on high- vs. low-performing programs via a cross-entropy update, and use their likelihood ratio to guide parent sampling.}
\label{fig:algo_concept_trees}
\end{figure}

To implement the ideas above, we introduce Contrastive Concept-Tree Search (CCTS), a concept-guided algorithm-discovery framework that learns which semantic concepts to prioritize during search. CCTS uses a cross-entropy update in concept space and a Tree-structured Parzen Estimator (TPE) to contrast high- and low-performing algorithms, producing a likelihood-ratio score that guides parent selection and semantic edits toward concept combinations associated with improvement (Fig.~\ref{fig:algo_concept_trees}).

The paper is organized as follows. Sec.~\ref{sec:related} reviews related work on LLM-assisted algorithm discovery and evolutionary search. Sec.~\ref{sec:method} introduces the general evolutionary algorithm framework for algorithm discovery that underlies many existing approaches. Sec.~\ref{sec:CCTS} then presents Contrastive Concept-Tree Search in detail. Sec.~\ref{sec:experiments} reports numerical results and empirical comparisons. For clarity, the notation used throughout this paper is summarized in App.~\ref{sec:notations}.

\section{Related works \label{sec:related}}


FunSearch \cite{romera2024mathematical} is one of the first widely recognized methods to reignite interest in evolutionary approaches to algorithm discovery by combining them with pre-trained LLMs, using the model as a mutation operator within an island-based search loop. Building on this idea, AlphaEvolve \cite{novikov2025alphaevolve} generalizes the framework into a more modular and robust system applicable across domains, including mathematical discovery \cite{georgiev2025mathematical} and, in particular, combinatorics. More recent results on SAT solver design further show that this paradigm can outperform highly optimized human-engineered systems \cite{yu2025autonomous}.

Several recent works extend the evolutionary algorithm-discovery by modifying candidate generation or evaluation. Hercules \cite{wu2025efficient} extracts reusable abstractions from strong heuristics and reinjects them into prompts, LASR \cite{grayeli2024symbolic} abstracts semantic patterns from high-performing programs to guide LLM-based mutation, while ReEvo \cite{ye2024reevo} uses LLM-generated reflections over performance differences to guide subsequent mutations. Evolution of Heuristics (EoH) \cite{liu2024evolution} evolves both reasoning traces and executable code, and HeurAgenix \cite{yang2025heuragenix} adopts a two-stage hyper-heuristic strategy that separates heuristic discovery from selection. ThetaEvolve \cite{wang2025thetaevolve} further augments an AlphaEvolve-style loop with test-time reinforcement learning to adapt the LLM’s generation policy online.



Despite these variations, most LLM-assisted algorithm-discovery systems are based on evolutionary algorithms as the outer search loop, reflecting the highly irregular structure of program space. Programs are modified through discrete mutations rather than smooth interpolation, selection is driven by fitness, and diversity is typically maintained via islands, niches\cite{hu2025partition}, or MAP-Elites-style archives \cite{novikov2025alphaevolve,mouret2015illuminating}. Because algorithm space lacks a simple or well-behaved structure, principled guided search methods such as estimation-of-distribution approaches have remained relatively underexplored.

The algorithm-discovery setting considered here is related to prompt-engineering methods, including evolutionary prompt search \cite{guo2025evoprompt}, tree-based reasoning \cite{yao2023tree}, Bayesian optimization over prompts or reasoning traces \cite{schneider2024hyperband}, and other black-box prompt optimization techniques \cite{pryzant2023automatic}. A key difference is that algorithm discovery typically assumes a well-defined, externally computable evaluator, enabling direct use of empirical performance signals. For simplicity, we keep prompt engineering minimal and treat the LLM as a black-box generator, focusing instead on principled parent selection and search dynamics over algorithm space.

\section{Problem Setup}
\label{sec:method}

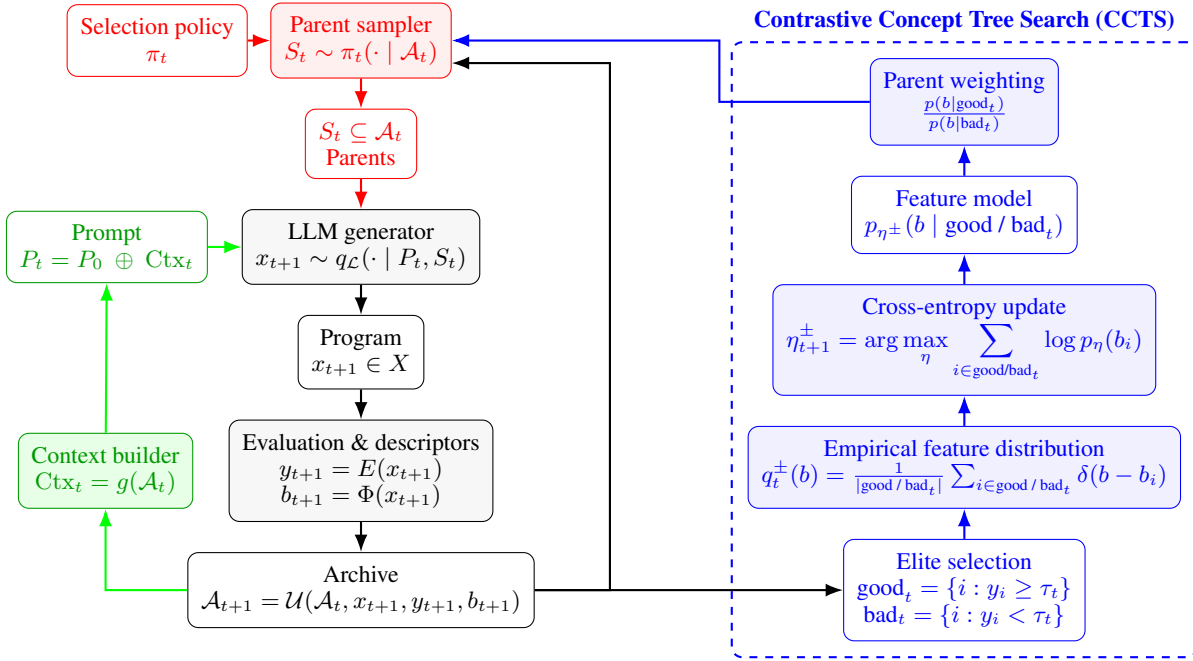
\begin{figure*}[htbp]
\centering
\begin{tikzpicture}[
    font=\small,
    >=Latex,
    node distance=4mm and 3mm,
    box/.style={draw, rounded corners, align=center, inner sep=5pt},
    op/.style={draw, rounded corners, align=center, inner sep=5pt, fill=black!3},
    addbox/.style={draw=blue, text=blue, rounded corners, align=center, inner sep=5pt},
    addop/.style={draw=blue, text=blue, rounded corners, align=center, inner sep=5pt, fill=blue!6},
    addarrow/.style={->, draw=blue},
    addarrowthick/.style={->, thick, draw=blue},
    adddashed/.style={draw=blue, dashed, rounded corners, inner sep=6pt},
    selbox/.style={draw=red, text=red, rounded corners, align=center, inner sep=5pt},
    selop/.style={draw=red, text=red, rounded corners, align=center, inner sep=5pt, fill=red!6},
    ctxbox/.style={draw=green!60!black, text=green!60!black,
               rounded corners, align=center, inner sep=5pt},
    ctxop/.style={draw=green!60!black, text=green!60!black,
              rounded corners, align=center, inner sep=5pt, fill=green!10}
]


\node[selop] (sampler)
{Parent sampler\\
$S_t \sim \pi_t(\cdot\mid\mathcal{A}_t)$};

\node[selbox, left=of sampler] (policy)
{Selection policy\\
$\pi_t$};

\node[selbox, below=of sampler] (parents)
{$S_t \subseteq \mathcal{A}_t$\\Parents};

\node[op, below=of parents] (gen)
{LLM generator\\
$x_{t+1} \sim q_{\mathcal{L}}(\cdot\mid P_t,S_t)$};

\node[box, below=of gen] (x)
{Program\\$x_{t+1}\in X$};

\node[op, below=of x] (eval)
{Evaluation \& descriptors\\
$y_{t+1}=E(x_{t+1})$\\
$b_{t+1}=\Phi(x_{t+1})$};

\node[box, below=of eval] (A)
{Archive\\$\mathcal{A}_{t+1}=\mathcal{U}(\mathcal{A}_t,x_{t+1},y_{t+1},b_{t+1})$};

\node[ctxop, left=of eval, xshift=-1.7mm] (ctx)
{Context builder\\
$\mathrm{Ctx}_t = g(\mathcal{A}_t)$};

\node[ctxbox, left=of gen, xshift=-1.5mm] (prompt)
{Prompt\\$P_t = P_0\ \oplus\ \mathrm{Ctx}_t$};

\draw[->, draw=red, thick] (policy) -- (sampler);
\draw[->, draw=red, thick] (sampler) -- (parents);
\draw[->, draw=red, thick] (parents) -- (gen);

\draw[->, thick] (gen) -- (x);
\draw[->, thick] (x) -- (eval);
\draw[->, thick] (eval) -- (A);

\draw[->, draw=green, thick] (ctx) -- (prompt);
\draw[->, draw=green, thick] (prompt) -- (gen);

\draw[->, draw=green, thick] (A.west) -| (ctx.south);


\node[addbox, right=of A, xshift=38mm] (elite)
{Elite selection\\
$\text{good}_t=\{i:y_i\ge\tau_t\}$\\
$\text{bad}_t=\{i:y_i<\tau_t\}$};

\node[addop, above=of elite] (empirical)
{Empirical feature distribution\\
$q^{\pm}_t(b)=\frac{1}{|\text{good / bad}_t|}\sum_{i\in\text{good / bad}_t}\delta(b-b_i)$};

\node[addop, above=of empirical] (ce)
{Cross-entropy update\\
$\displaystyle
\eta^{\pm}_{t+1}
=
\arg\max_\eta
\sum_{i\in\text{good/bad}_t}
\log p_\eta(b_i)
$};

\node[addbox, above=of ce] (model)
{Feature model\\
$p_{\eta^{\pm}}(b\mid\text{good / bad}_t)$};

\node[addop, above=of model] (weight)
{Parent weighting\\
$\frac{p(b\mid\text{good}_t)}{p(b\mid\text{bad}_t)}$};

\draw[addarrow, thick] (elite) -- (empirical);
\draw[addarrow, thick] (empirical) -- (ce);
\draw[addarrow, thick] (ce) -- (model);
\draw[addarrow, thick] (model) -- (weight);

\draw[addarrow, draw=black, thick] (A.east) -- (elite.west);

\draw[addarrowthick, thick]
  (weight.west)
  -- ++(-20mm,0)
  |- (sampler.east);

\draw[->, thick]
  (A.east)
  -- ++(10mm,0)
  |- ([yshift=-3mm]sampler.east);

\node[adddashed, thick, fit=(elite)(empirical)(ce)(model)(weight),
      label={[text=blue]above:{\textbf{Contrastive Concept Tree Search (CCTS)}}}] {};

\end{tikzpicture}
\caption{Schematic of the LLM-assisted evolutionary search loop. The red blocks implement the parent selection process, the grey blocks implement the child generation, and the green blocks describe the prompt update process. The blue blocks describe the main novelty of this work: a contrastive exploration process in context space that is used to inform the parent generation process. Boxes containing operators and objects are denoted by dark and light colors, respectively. The notation appearing in the blocks is defined in Sec. \ref{sec:method} and Sec. \ref{sec:CCTS}.}
\label{fig:llm_evo_schematic_guided}
\end{figure*}

We formalize here a general framework for algorithm discovery that encompasses previous approaches and serves as a foundation for our method, which is schematically summarized by the diagram in Fig.~\ref{fig:llm_evo_schematic_guided}.


\subsection{Search spaces}

We consider tasks defined by a formal problem specification and a verifier or evaluator. Concretely, a task $C$ is characterized by (i) an external evaluation function $E$ that takes an algorithm $x$ and returns a scalar score measuring performance, and (ii) a natural-language description $P_0(C)$ that specifies the objective, constraints, or rules of the problem (e.g., a mathematical problem statement). The evaluator $E$ allows ground-truth comparison of candidate algorithms.

Candidate algorithms are represented as runnable programs $x \in X$, where $X$ denotes the genotype (code) space, such as source code, pseudocode, or structured algorithmic descriptions. Programs are generated conditionally by a large language model (LLM), which implicitly restricts search to a learned subset $\mathcal{W} \subset X$ corresponding to the LLM’s internal representation of plausible algorithms.

Each program $x$ is evaluated to produce a fitness value $y = E(x; C)$, and may also be mapped to a feature or phenotype representation $b = \Phi(x)$, which captures behavioral or semantic properties of the algorithm. These features are used for selection, diversity maintenance, or analysis, and may themselves be extracted from the LLM, by directly prompting it to generate a list of relevant concepts pertaining to a generated program. The overall search dynamics are further controlled by a set of hyperparameters $\theta$, which govern aspects such as population size, selection policies, and generation settings. The list of important hyperparameters is summarized in App. Table \ref{tab:operators_and_hyper}.

The objective of algorithm discovery is to identify an algorithm that maximizes performance on the task,
\begin{align}
x^\star \in \arg\max_{x \in X} \; E(x; C),
\end{align}
using only black-box access to the evaluator and the conditional generation capabilities of the LLM.

\subsection{Evolutionary search loop}

Algorithm discovery proceeds iteratively through an evolutionary loop. At iteration $t$, the algorithm maintains an archive $\mathcal{A}_t = \{(x_i, y_i, b_i)\}_{i=1}^{N_t}$ containing previously evaluated programs along with their fitness and features.

Parent selection is defined by a sampling distribution $\pi_t(\cdot \mid \mathcal{A}_t)$ over subsets of the archive. Selected parents $S_t \subseteq \mathcal{A}_t$ are used to generate a new child program by conditioning an LLM on both the parents and a prompt. Formally, letting $\mathcal{L}$ denote the LLM, a child program $x_{t+1}$ is sampled via
\begin{align}
x_{t+1} \sim q_{\mathcal{L}}(\cdot \mid P_t, S_t),
\end{align}
where $P_t$ is a prompt constructed from the task description and additional context and $q_{\mathcal{L}}$ denotes LLM sampling.

The prompt is decomposed into a fixed task prompt $P_0(C)$ and an adaptive context,
\begin{align}
P_t = P_0(C) \oplus \mathrm{Ctx}_t,
\end{align}
where $\mathrm{Ctx}_t = g(\mathcal{A}_t, S_t; \mathcal{L})$ summarizes information from the current archive. This context may include descriptions of recent improvements, differences between parent and child programs, or feedback from failed evaluations. In practice, such adaptive prompting encourages local refinement within the LLM-induced program manifold.

Given the child, we evaluate its fitness and features as 
\begin{align}
y_{t+1} = E(x_{t+1}; C), \qquad b_{t+1} = \Phi(x_{t+1}),
\end{align}
and the archive is updated accordingly.

\subsection{Parent selection policies\label{sec:parent_selection}}

Parent selection plays a central role in shaping the search dynamics. A general parent-selection policy is defined as a probability distribution over subsets of the archive,
\begin{align}
S_t \sim \pi_t(\cdot \mid \mathcal{A}_t),
\end{align}
which may depend on fitness values, extracted features, or other archive statistics.

In practice, parent selection is often implemented as a mixture of simple strategies. In this paper, we use the following strategies:
\begin{itemize}
    \item \textbf{Uniform selection}: parents are sampled uniformly over the archive (or current island), providing an exploration baseline that is independent of fitness or features.
    
    \item \textbf{Greedy selection}: the highest-scoring algorithm in the archive is selected as the parent, corresponding to pure exploitation based on fitness.
    
    \item \textbf{$k$-elite selection} \cite{zames1981genetic}: parents are sampled uniformly from the top-$k$ algorithms ranked by fitness, interpolating between uniform exploration and greedy exploitation.
    
    \item \textbf{CCTS (this work)}: parents are sampled according to a learned, feature-based weighting that exploits semantic information extracted from programs; this strategy is described in detail in the next section.
\end{itemize}

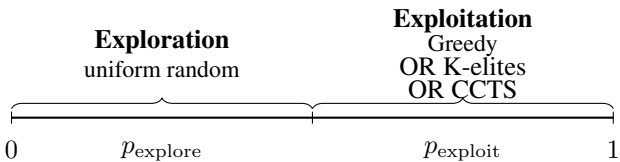
\begin{figure}[htbp]
\centering


\begin{tikzpicture}[x=8cm,y=1cm]

  \draw[thick] (0,0) -- (1,0);
  \draw (0,0.08) -- (0,-0.08) node[below=3pt] {$0$};
  \draw (1,0.08) -- (1,-0.08) node[below=3pt] {$1$};

  \draw (0.5,0.08) -- (0.5,-0.08);

  \node[align=center] at (0.25, 0.85)
    {\textbf{Exploration}\\[-1pt]\footnotesize uniform random};

  \node[align=center] at (0.75, 0.85)
    {\textbf{Exploitation}\\[-3pt]\footnotesize
      Greedy\\[-3pt]
      OR K-elites\\[-3pt]
      OR CCTS};

  \node at (0.25, -0.45) {$p_{\mathrm{explore}}$};
  \node at (0.75, -0.45) {$p_{\mathrm{exploit}}$};

  \draw[decorate,decoration={brace,amplitude=4pt}]
    (0,0.12) -- (0.5,0.12);
  \draw[decorate,decoration={brace,amplitude=4pt}]
    (0.5,0.12) -- (1,0.12);

\end{tikzpicture}

\caption{Schematic description of the stochastic parent-selection policy mixing uniform exploration and exploitation. At each step, a parent can be selected either uniformly from the archive, with exploration probability $p_\text{explore}$ (exploration), or it can be selected according to a given performance-informed strategy with probability $p_\text{exploit}=1-p_\text{explore}$. Segment sizes are schematic.}
\label{fig:mixture_selection_simple}
\end{figure}

In this paper, we consider a two-component parent-selection mixture that balances exploration via uniform sampling with exploitation using one of the remaining strategies. Figure~\ref{fig:mixture_selection_simple} illustrates the resulting exploration–exploitation trade-off. 

\section{Method: Contrastive Concept-Tree Search\label{sec:CCTS}}

CCTS augments the general evolutionary loop of Section~\ref{sec:method} with an explicit semantic feature space and a lightweight estimation-of-distribution model \cite{pugh2016quality} that biases parent selection toward empirically useful regions of that space. The key design choice is to learn guidance in an interpretable concept space, which is then exploited to select the programs to mutate. 

\subsection{Concept representation and concept tree}

Each candidate program $x$ is mapped to a semantic feature representation $b=\Phi(x)$ extracted from the program by a direct prompt to the LLM after the generation process. In CCTS, $b$ is a set of activated concepts organized in a rooted concept tree. Concretely, let $V$ denote the set of concepts discovered so far. Each node $v\in V$ corresponds to a semantic concept, and edges encode refinement relations: a child concept $v$ is a more specific version of its parent, denoted by $\mathrm{pa}(v)$.

We represent the extracted concept set as an indicator vector $b=(b_v)_{v\in V}$ with $b_v\in\{0,1\}$. Concept activations are assumed to be ancestor-closed, $b_v = 1 \;\Rightarrow\; b_{\mathrm{pa}(v)} = 1$, $\forall v\neq \text{root}$, so that activating a refined concept implies activating all its ancestors. The concept tree is dynamic: as the search proceeds, the feature extractor may introduce new concepts (new nodes) that are inserted into the hierarchy (see Fig. \ref{fig:algo_concept_trees}). We say a concept is discovered once it appears in $\Phi(x)$ for some evaluated program and is added to $V$.

\subsection{Contrastive guided parent selection\label{sec:CCTS_contrastive}}

CCTS learns which concepts are empirically associated with improved performance in a contrastive manner, and uses this information to bias parent selection (see Fig. \ref{fig:llm_evo_schematic_guided}). We partition the archive of programs into \emph{good} and \emph{bad} subsets using a threshold $\tau_t$:
\begin{align}
\text{good}_t=\{i:y_i\ge \tau_t\},
\qquad
\text{bad}_t=\{i:y_i< \tau_t\}.
\end{align}

We then fit two probabilistic models over feature vectors, one for each subset:
\begin{align}
\hat p_{\eta^+}(b) \approx p(b\mid \text{good}_t),
\qquad
\hat p_{\eta^-}(b) \approx p(b\mid \text{bad}_t),
\end{align}
within a shared parametric family $\{\hat p_\eta\}_{\eta\in\Theta}$. Parameters are estimated by a cross-entropy update (equivalently maximum likelihood) on the corresponding subset:
\begin{align}
\eta_{t+1}^{\pm} \in \arg\max_\eta \sum_{i\in \text{good/bad}_t} \log \hat p_\eta(b_i),
\end{align}
with standard smoothing and regularization for numerical stability. Here, \( b_i = \Phi(x_i) \) is the concept vector of algorithm \( x_i \), and the sum is taken over the archive elements in the good or bad partition.

We bias parent selection using the likelihood-ratio weight
\begin{equation}
w(b_i)=\frac{\hat p_{\eta^+}(b_i)}{\hat p_{\eta^-}(b_i)},
\qquad
\pi_{\text{CCTS}}(x_i\mid \mathcal{A}_t) \propto w(b_i).
\label{eq:ratio_weight}
\end{equation}
This contrastive formulation emphasizes features that distinguish high- and low-performing programs.

Thus, the model operates entirely in feature space and guides search by reallocating sampling probability toward existing archive elements whose features are empirically associated with improvement. In practice, we define paired empirical feature distributions over high- and low-performance algorithms,
\begin{align}
q_t^{+}(b)
& \propto
\sum_{i \in \text{good}_t}
\delta(b - b_i),
\quad
q_t^{-}(b)
\propto \sum_{i \in \text{bad}_t}
\delta(b - b_i),
\end{align}
which represent the empirical distributions of features observed in the good and bad subsets of the archive.

In practice, the cross-entropy (CE) method \cite{rubinstein2004cross} is used to fit a parametric family
$\{\hat p_\eta(b)\}$ to each empirical distribution by minimizing the Kullback--Leibler divergence between the empirical measure and the model (see App. \ref{sec:fac_MLE_appendix} for details),
\begin{align}
\eta^{\pm}_{t+1}
&=
\arg\min_{\eta}
\mathrm{KL}\!\left(
q_t^{\pm}(b)
\;\middle\|\;
\hat p_\eta(b)
\right),
\end{align}
where the last expression corresponds to maximum likelihood estimation on samples drawn from
$q_t^{+}$ or $q_t^{-}$ respectively. Guided parent selection of a parent with concept vector $b$ relies then on the log-likelihood ratio $\log(w(b)) = \log \hat p_{\eta^+}(b) - \log \hat p_{\eta^-}(b)$, called log concept utility, to score archive elements.

\subsection{Hierarchical factorized and leaf-restricted model}

In our main implementation, we instantiate the contrastive feature model using a simple hierarchical, factorized distribution consistent with the concept-tree structure. This factorized formulation is closely related to Tree-structured Parzen Estimator (TPE)–style likelihood ratios over concept configurations \cite{bergstra2011algorithms}, and is computationally efficient, interpretable at the level of individual concepts, and sufficient to capture concept utility. Full details of the model and estimators are provided in Appendix~\ref{sec:fac_MLE_appendix}. When guiding child generation, we restrict explicit concept-level interventions to leaf nodes of the current concept tree.

\subsection{Exploration mechanism over new and rare concepts}

To avoid premature convergence and to encourage discovery of new or rarely explored concepts, we complement this exploitation strategy with a lightweight concept-level exploration mechanism operating directly on the concept tree. At a high level, this mechanism ensures nonzero support for newly discovered concepts, introduces a novelty bias favoring underexplored concepts, and combines exploitation and exploration through a simple mixture rule when selecting concepts to emphasize during generation. Concept selection is implemented at the prompt level by injecting the chosen concept as a semantic directive for the LLM. Full details of this exploration mechanism are provided in Appendix~\ref{sec:concept_exploration}. This concept exploration mechanism is used regardless of the parent sampling strategy for exploitation.

\section{Experiments \label{sec:experiments}}

To demonstrate the generality of the proposed method, we apply our framework to a benchmark constructed for this work and adapted from previous studies \cite{novikov2025alphaevolve, georgiev2025mathematical}, consisting of several mathematical problems drawn primarily from Erdős-style combinatorics, including the circle packing problem, the Arithmetic Kakeya conjecture, Heilbronn’s triangle problem, and square in square problem (see Appendix \ref{sec:benchmark} for details).


\subsection{Ablating core components}

\begin{figure*}[h]
    \centering
    \begin{subfigure}[t]{0.49\textwidth}
        \centering
        \caption{Real task: circle packing}
        \includegraphics[width=\textwidth]{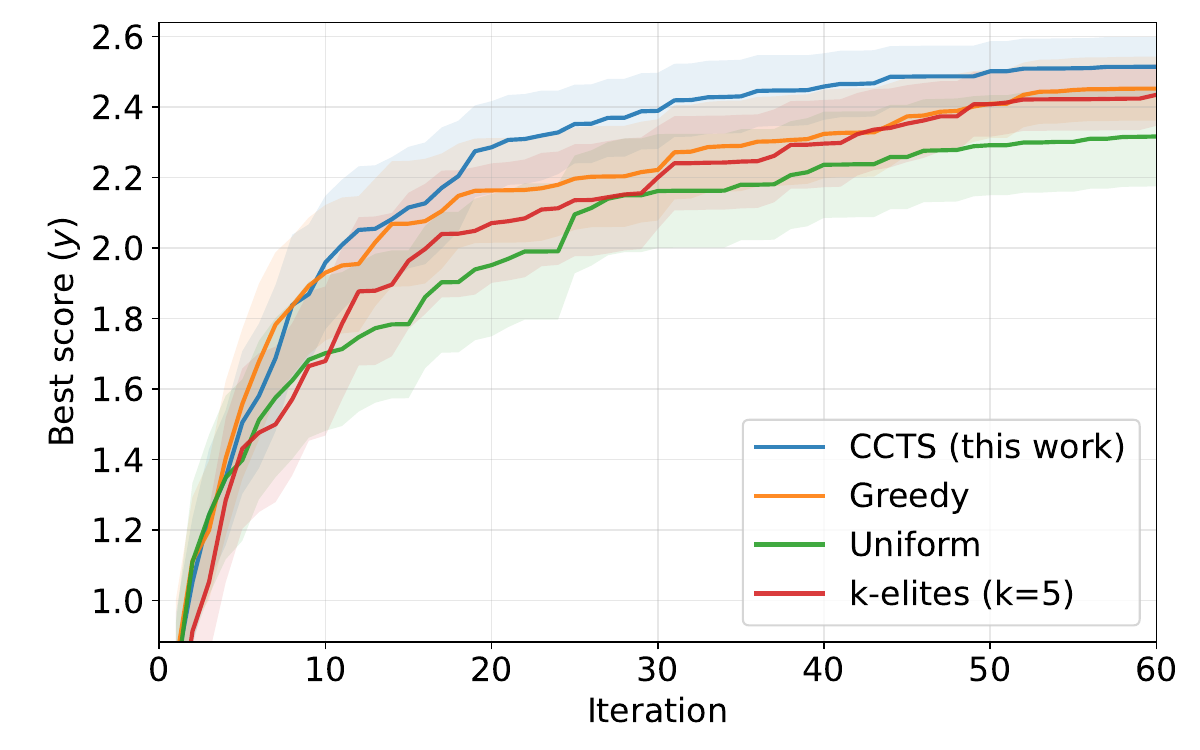}
    \end{subfigure}\hfill
    \begin{subfigure}[t]{0.49\textwidth}
        \centering
        \caption{Synthetic task}
        \includegraphics[width=\textwidth]{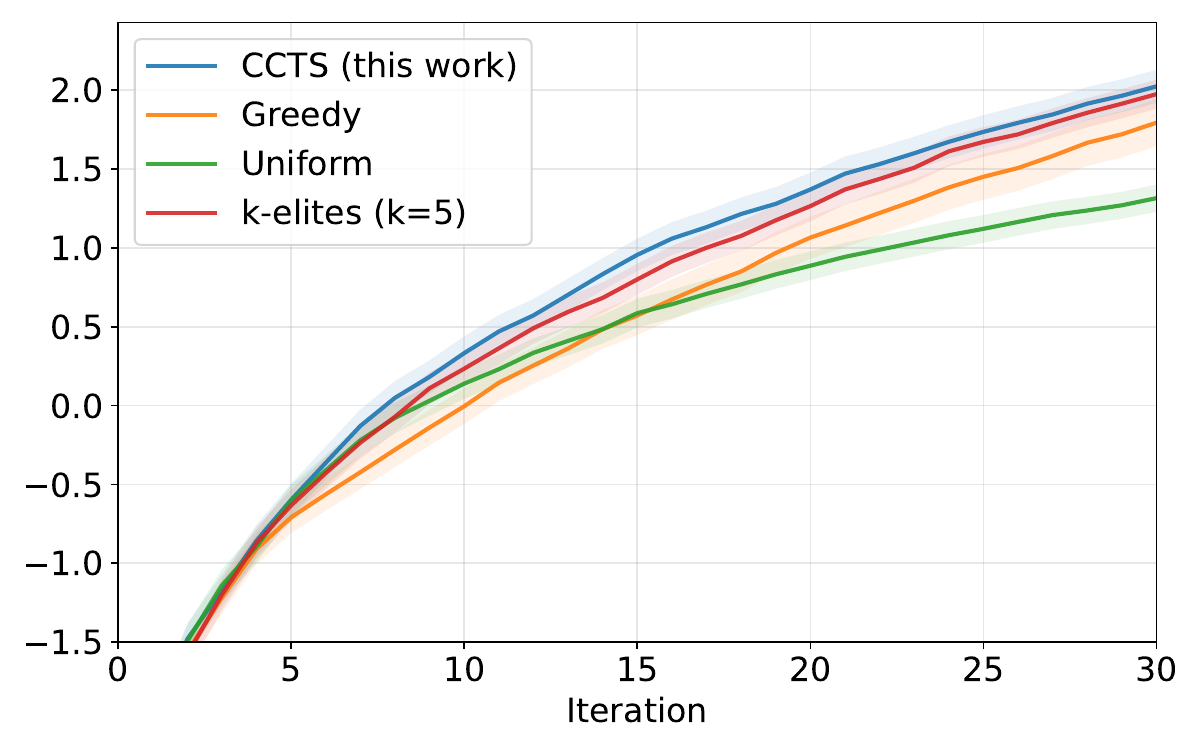}
    \end{subfigure}
    \caption{Best score vs.\ iteration (number of LLM calls) for the circle packing task. Ablations are described in the text. For all tasks, the probability of exploitation is set to $p_{\text{exploit}} = 0.85$. (a) Ran with gemini-flash-2.0 and averaged over 60 runs. (b) Synthetic task averaged over 500 runs. Shaded area show the 95\% confidence interval.}
    \label{fig:ablation_core}
\end{figure*}

We compare CCTS against standard parent-selection baselines defined in Section~\ref{sec:method}.
The full setting (“CCTS”) uses concept-guided parent selection together with concept exploration and LLM-based local mutation.
As baselines, we consider Greedy, $k$-elites ($k=5$), and Uniform parent selection strategies (see Section~\ref{sec:method}).
For methods that alternate between exploitation and exploration (CCTS, Greedy, and $k$-elites), the exploitation strategy is selected with probability $p_{\text{exploit}} = 0.85$ (see Fig. \ref{fig:mixture_selection_simple}), and uniform sampling is used otherwise. As shown in Fig.~\ref{fig:ablation_core}(a), CCTS achieves higher scores for any fixed number of iterations, thus demonstrating the benefit of exploiting the structure of the concept space.

\begin{figure*}[h]
    \centering
    \includegraphics[width=0.8\textwidth]{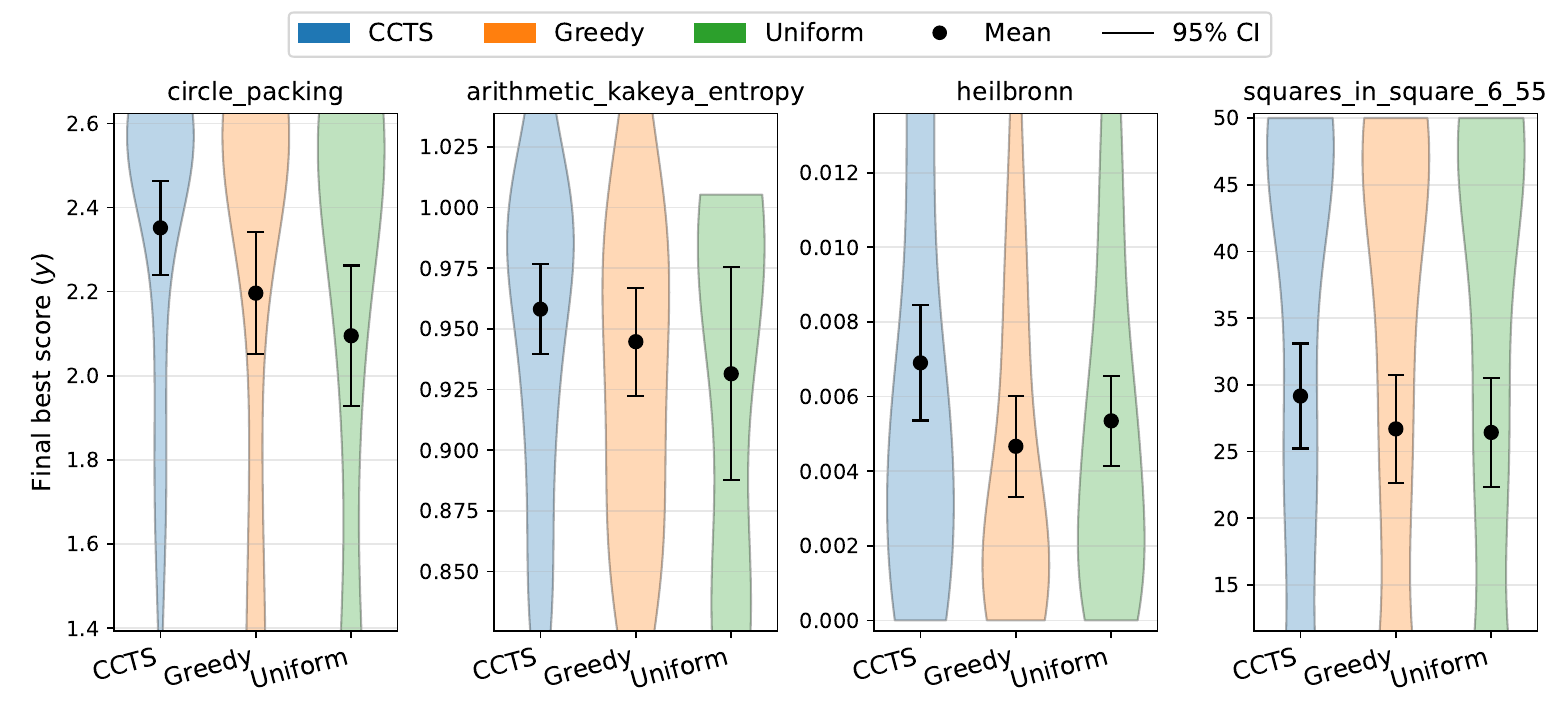}
    \caption{Performance of CCTS compared to baseline methods across multiple tasks. Violin plots show the distribution of the best score after 25 iterations, aggregated over 60 runs; the mean and 95\% confidence interval are superimposed. All tasks are run with gemini-flash-2.0 and $p_{\text{exploit}} = 0.85$. The distribution of best scores for CCTS presents thinner tails at low values of final best scores.}
    \label{fig:ablation_summary}
\end{figure*}

To assess the robustness of CCTS' edge over the baseline methods, we compare the best score achieved by CCTS over several runs against the one achieved through other methods, across multiple mathematical tasks. Our results are displayed in Fig.~\ref{fig:ablation_summary}, and they demonstrate that CCTS consistently achieves higher scores and faster improvement than the baseline methods, despite substantial diversity in problem structure and evaluation criteria.

\subsection{Comparisons in synthetic algorithm discovery task}

To analyze our results, we construct a synthetic algorithm-discovery environment in which the LLM is replaced by a simplified hand-crafted operator. This controlled setting allows us to study the learning dynamics of CCTS in isolation and to verify that the proposed approach yields computational advantages independent of language-model-specific effects. In this synthetic model, a ground-truth (teacher) concept tree is generated through a stochastic branching process with base branching ratio $\lambda_0$, and a latent ground-truth utility, drawn from a Gaussian distribution, is assigned to each concept. An algorithm is then identified by a set of active concepts on the tree, and its quality is determined by the sum of the utilities of its active concepts. Child generation replaces the LLM with a simple mutation operator that performs local and global edits on the parent’s concepts, guided by the teacher concept tree. See App.~\ref{sec:synthetic} for details of the synthetic construction and operators. Beyond validating the generality of our finding, this model enables the formulation of testable hypotheses about how properties of latent concept spaces influence the algorithm-discovery dynamics.

Fig. \ref{fig:ablation_core} (b) shows the score achieved by the best algorithm upon several iteration of the automatic search in the synthetic environment, under ablations that are functionally analogous to those performed in the experiments. The resulting curves closely mirror those observed with the LLM-based experiments, suggesting that the simple synthetic model captures key qualitative aspects of the search dynamics induced by LLM-assisted algorithm discovery.

Crucially, the synthetic model allows us to compare the learned concept utilities to the synthetic ground truth.
Fig.~\ref{fig:comparison_W} shows strong agreement between the teacher and student concept weights, indicating that CCTS recovers both the structure and relative utility of latent concepts. This result confirms that CCTS learns meaningful semantic representations that can effectively guide algorithm discovery.

\subsection{Extracted concepts}

\begin{figure*}[t]
\centering
\begin{minipage}[t]{0.35\linewidth}
    \vspace{0pt}\raggedright 
    \begin{subfigure}[t]{\linewidth}
        \centering
        \caption*{Inferred concept utility (Circle packing)}
        \vspace{-0.03in}
        \includegraphics[width=1.0\linewidth]{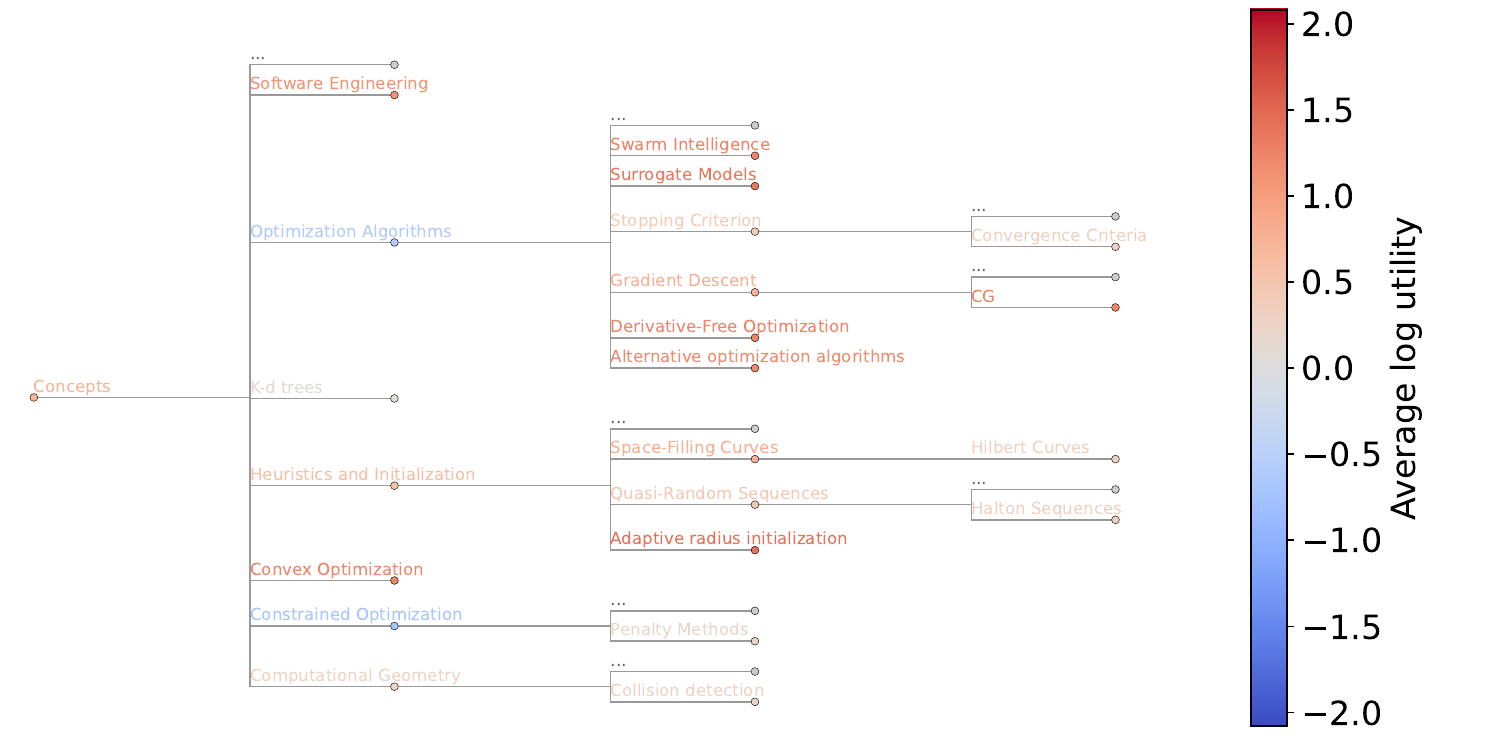}
        \label{fig:LLM_task_tree}
    \end{subfigure}
\end{minipage}
\begin{minipage}[t]{0.35\linewidth}
    \vspace{0pt}\raggedright 
    \begin{subfigure}[t]{\linewidth}
        \centering
        \caption*{Inferred concept utility (Heilbronn triangle)}
        \vspace{-0.03in}
        \includegraphics[width=1.0\linewidth]{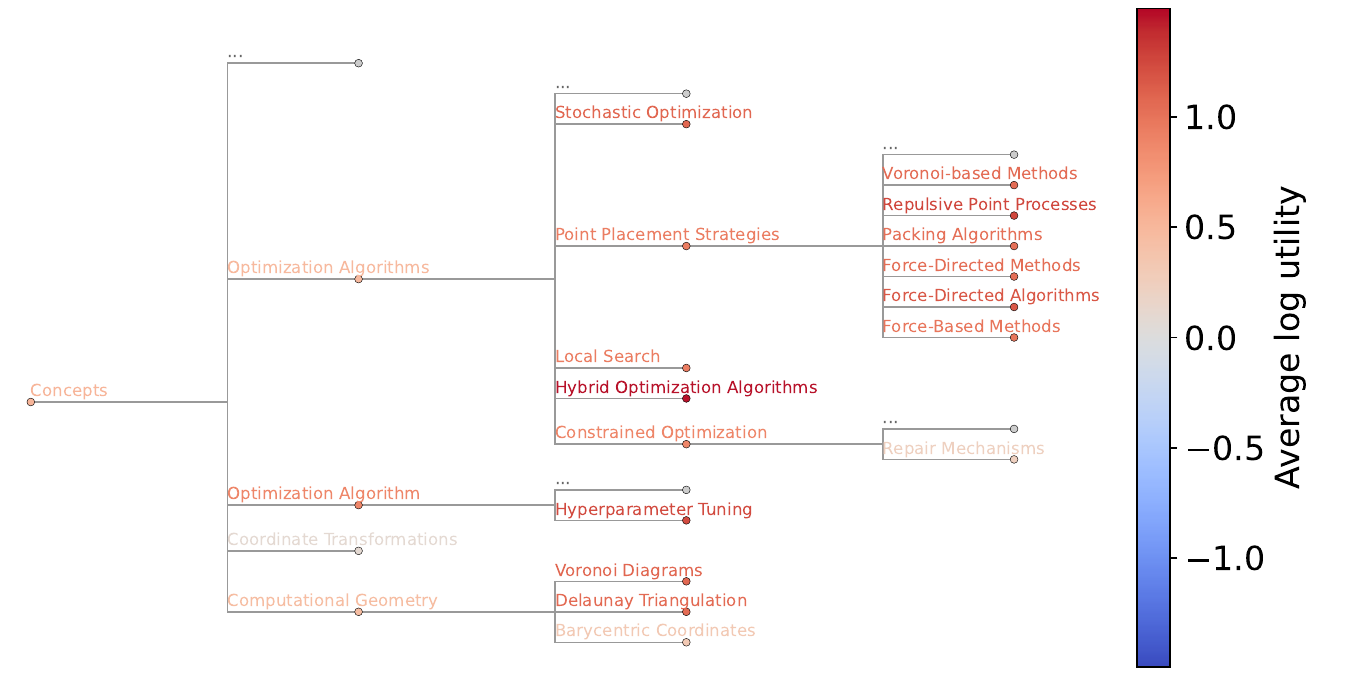}
        \label{fig:synthetic_task_tree}
    \end{subfigure}
\end{minipage}
\begin{minipage}[t]{0.28\linewidth}
    \vspace{0pt}\raggedright 
    \begin{subfigure}[t]{\linewidth}
        \centering
        \includegraphics[width=0.9\linewidth]{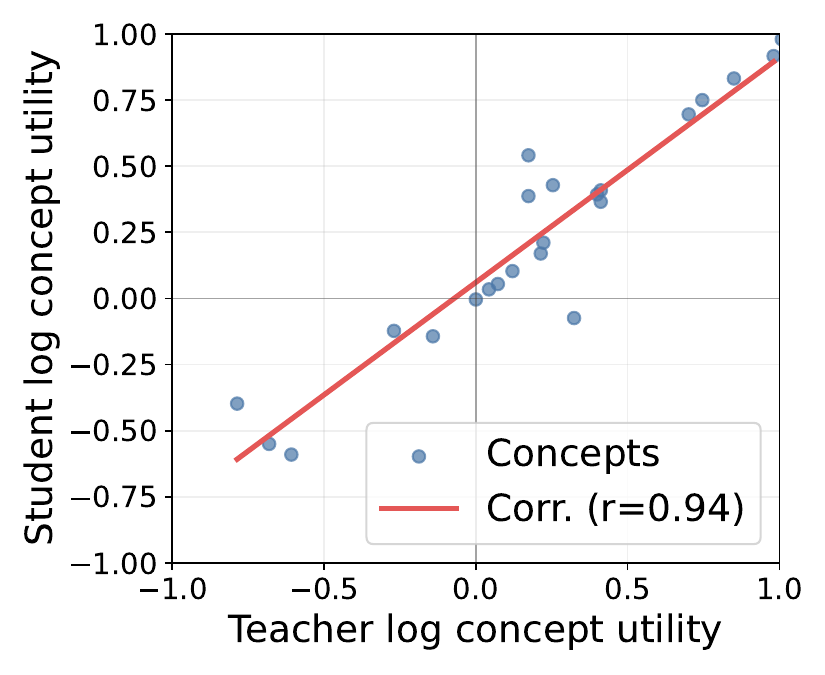}
        \label{fig:synthetic_task}
    \end{subfigure}
\end{minipage}
\caption{Concept trees extracted by the LLM for the circle packing (left) and Heilbronn triangle (middle) tasks. Tree nodes corresponding to the top and bottom 10 inferred concept utilities are shown, with warm and cool colors indicating useful and non-useful concepts, respectively. Utility of concepts are averaged over 60 runs. See App. \ref{sec:interpretability} for details. (Right) Correlation between teacher and student log concept utility, $\log w$, for the synthetic task (see App.~\ref{sec:synthetic} for details).}
\label{fig:comparison_W}
\end{figure*}

We next analyze the semantic concepts extracted by the LLM during search. The concept trees obtained for two real tasks are shown in Fig.~\ref{fig:comparison_W}. The resulting concepts are meaningfully related to the underlying problem, capturing both core structural elements of circle packing and refined, higher-level algorithmic strategies. Their hierarchical organization reflects increasing levels of specialization and provides an interpretable representation of the semantic structure explored during search. The growth in the number of discovered and assigned concepts over iterations is shown in App. Fig.~\ref{fig:concept_growth}.

\begin{figure}[h]
    \centering
    \includegraphics[width=0.5\textwidth]{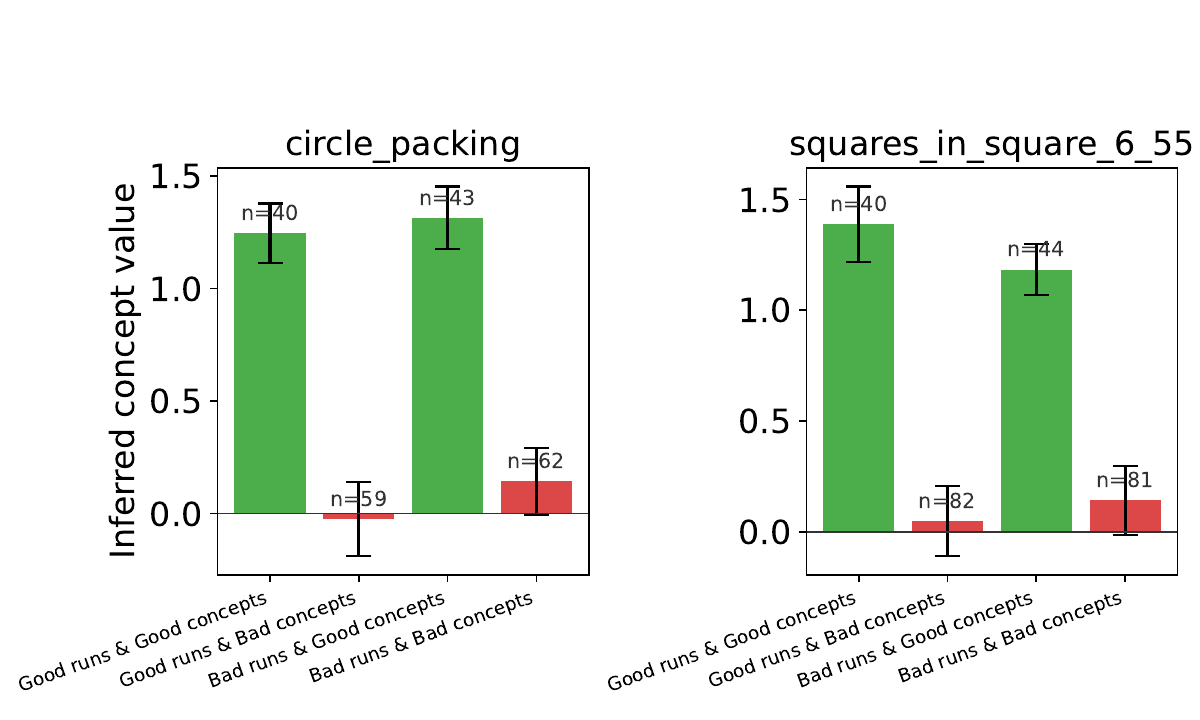}
    \caption{Inferred log concept utility $\log(w)$ , averaged over the four quadrants defined by run performance (``Good'' and ``Bad'' runs are the ones with top-50\% and bottom-50\% final-score) and concept utility rank (``Good'' and ``Bad'' concepts are the top-50\% and bottom-50\% ones ranked by $\log(w)$). Each value is averaged over $n$ concept–run instances.}
    \label{fig:log_lift_summary_quadrant_log_lift}
\end{figure}

What drives CCTS’s performance gains?  To answer this question, we rank runs by their final scores (high-performing versus low-performing runs) and rank concepts $b$ according to their utility, measured by the log-weight $\log w(b)$ defined in Eq.~(\ref{eq:ratio_weight}). In Fig.~\ref{fig:log_lift_summary_quadrant_log_lift}, we compare high-performing and low-performing runs by averaging, across runs, the log-utility of the top and bottom $10$ concepts ranked by their average $\log w(b)$. We observe that both successful and unsuccessful runs identify a similar set of high-utility (good) concepts, whereas successful runs learn substantially more about low-utility (bad) concepts, suggesting that an effective search is driven primarily by learning which concepts to avoid, while also identifying useful semantic components. This idea rationalizes the result displayed in Fig.~\ref{fig:ablation_summary}, where we see that the improvement in average score under CCTS is associated with a reduced width of the lower tail of the performance distribution. For further qualitative detail, an interpretable concept tree with diverse extracted concepts relevant to the circle packing task is shown in App.~Fig.~\ref{fig:concept_tree_dendrogram_LLM}, and a heatmap detailing the inferred per-run concept utilities is shown in App.~Fig.~\ref{fig:log_lift_summary_circle} (see App.~\ref{sec:interpretability} for details).

LLMs encode many broad, task-agnostic correlations acquired during pretraining, some of which are probably irrelevant for the specific objective at hand. As a result, the LLM may initially propose algorithms that incorporate some misleading concept. The contrastive probabilistic model corrects this behavior by identifying concepts whose presence does not correlate with improvement under the task evaluation. In its current form, CCTS appears to rely primarily on suppressing task-specific spurious correlations, thereby stabilizing search and reducing unproductive exploration.

\subsection{Concept structure dependent strategy}

\begin{figure}[htbp]
    \centering
    \includegraphics[width=\columnwidth]{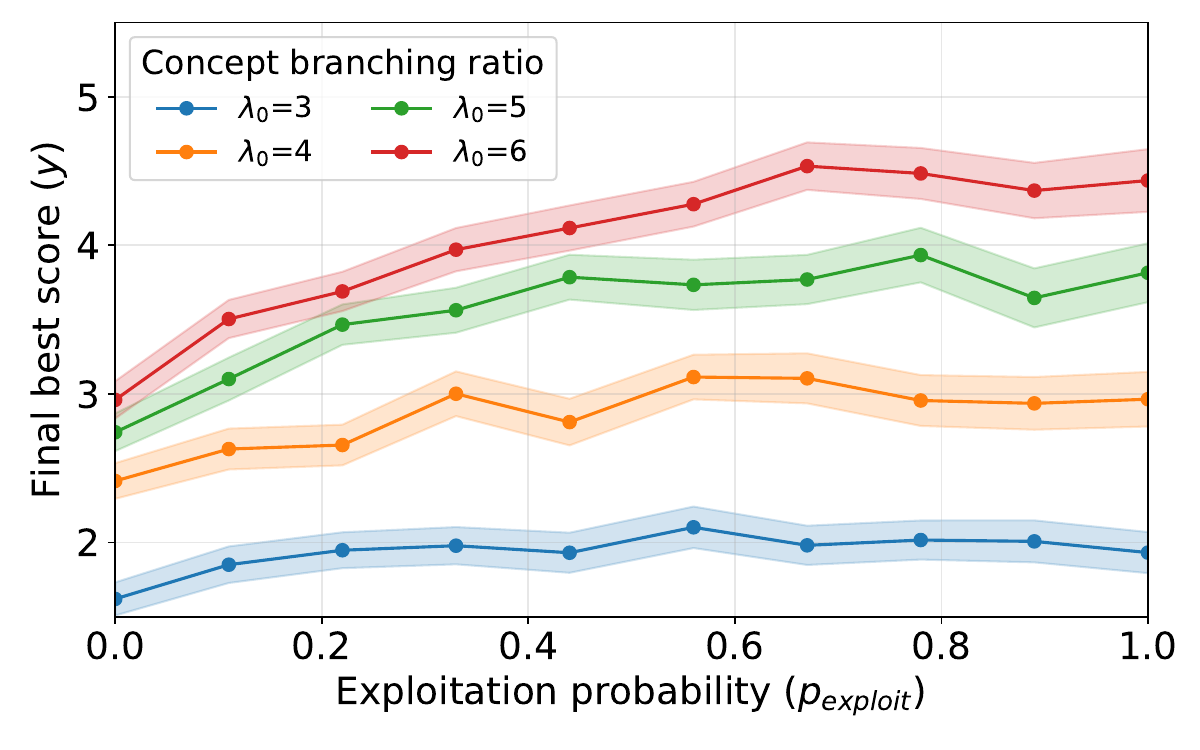}
    \caption{Best final score in synthetic task as a function of the exploitation probability for different branching ratios in synthetic tasks. The optimal search strategy of CCTS depends on the base branching ratio of the synthetic teacher concept tree.}
    \label{fig:exploit_vs_lambda}
\end{figure}

The synthetic model allows us to make predictions relating the optimal search strategy and the structure of the latent concept tree. Depending on the shape of the concept tree, the optimal strategy changes: exploitation is more effective for wider concept trees (with a higher branching ratio in the teacher concept tree), whereas exploration is preferable for narrower ones (see Fig.~\ref{fig:exploit_vs_lambda}).
When using an LLM on a real task, we observe that the optimal exploitation probability $p_{\text{exploit}}$ varies across problems and experimental settings (see Appendix~\ref{sec:exploit}). We conjecture that these variations may reflect differences in the underlying latent concept tree structure. Of course, further work is needed to clarify how properties of the latent concept tree influence the resulting search dynamics.

\section{Limitations and Broader Impact}

In this work, we showed that CCTS primarily improves the lower tail of the algorithms' score distribution. Strengthening guidance in the upper tail, for example by learning cross-concept correlations or enabling cross-mutation of multiple parents to better combine promising semantic components, remains an important direction for future work.

The benchmark employed in this work has the primary goal to allow the systematic study of the dynamics of algorithm discovery, while solving the underlying problems remained a secondary objective. 

Nevertheless, we report in App.~\ref{sec:circle_packing} that the solution found for the circle packing task reproduces the optimal result achieved by AlphaEvolve \cite{novikov2025alphaevolve}. The best solutions obtained for different problems will be addressed separately. Since our focus is to assessing whether and how CCTS accelerates improvement relative to baseline strategies, we restricted our experiments to relatively small, low-cost models (Gemini-2.0-Flash, etc.) and modest search budgets (see App. \ref{sec:choiceLLM} for a comparison between LLM models). In this work, our emphasis is on demonstrating relative, rather than absolute, performance improvements. Evaluating the potential of this approach for solving previously open or substantially harder problems will require scaling to larger models, longer runs, and more extensive compute.

We note that, although it is not the focus of this work, prompt engineering has a substantial impact on performance. In particular, insufficient task-specific information in the initial prompt can lead to overly generic concept trees, causing the search to stall as if trapped in a flat plateau of the algorithm space.

A possible performance improvement to be explored in the future consist in combining CCTS with additional population-structuring mechanisms that improve robustness and scalability but are largely orthogonal to the core search logic. Common examples include island-based evolution \cite{novikov2025alphaevolve}, and quality--diversity methods such as MAP-Elites \cite{novikov2025alphaevolve,mouret2015illuminating}, These features have not been explored here, in order to focus on the effect of parent sampling on the search, but their combination with CCTS  may further improve performance.

\section{Conclusion}

Fitness-based parent selection is a natural baseline in LLM-assisted algorithm discovery, but it is often a poor proxy for parent evolvability: high-scoring algorithms do not necessarily generate strong descendants. Contrastive Concept-Tree Search addresses this mismatch by shifting selection from raw fitness to semantic evidence of improvement, using contrastive concept statistics to guide search. By learning a task-conditioned semantic structure at runtime, CCTS allows search to operate in a concept space where distance, recombination, and search direction are meaningful, leading to significantly more efficient algorithm discovery.

\clearpage

\section*{Impact Statement}

This work studies automated algorithm-discovery systems inspired by multi-agent scientific discovery. A key societal challenge is ensuring interpretability and controllability as such systems scale.

By learning explicit, human-compatible semantic concept structures, the proposed approach aims to make automated discovery more transparent and easier to interact with and supervise. Such representations may help reduce risks and unintended harm by enabling clearer human oversight and intervention. The primary impact of this work is methodological and focused on improving the safety and interpretability of automated discovery systems.

\bibliography{arxivs}
\bibliographystyle{icml2026}


\appendix
\onecolumn

\section{Appendix}

\subsection{Notation \label{sec:notations}}

\subsubsection{Spaces and representations}

\begin{table}[htbp]
\centering
\caption{Key spaces involved in LLM-assisted evolutionary algorithm discovery.}
\begin{tabular}{l p{10cm}}
\textbf{Space} & \textbf{Role} \\
\hline

Prompt space $P$ 
    & Conditioning inputs that guide LLM-based program generation. \\
    
Genotype / code space $X$ 
    & Concrete algorithmic representations generated by the LLM. \\

LLM world-model space $\mathcal{W} \subset X$
    & LLM’s learned representation $x \in \mathcal{W}$ of the algorithmic manifold. \\
    
Phenotype / feature space $b=\Phi(x)$ 
    & Descriptor of algorithm behavior used for selection and diversity. \\

    
Hyperparameter space $\theta$
    & Hyperparameters of the evolutionary dynamics. \\

Fitness space $y=E(x;C)$
    & Performance signal used to evaluate and select algorithms. \\

\hline
\end{tabular}
\label{tab:space}
\end{table}

An important characteristic of the algorithm search space is that it is multi-layered, as detailed in Table~\ref{tab:space}. The central layer is the space of algorithms $x \in X$, or the genotype space, representing a program, a function, a file, or a repository patch. The LLM induces a constrained manifold
$\mathcal{W}$ over the genotype space $X$, within which new algorithms are generated
conditioned on prompts $P$. Each candidate $x\in X$ is evaluated to produce a fitness
$y$ and mapped to observable descriptors $b$, which are used to guide selection and
maintain diversity. External hyperparameters $\theta$ control the resulting evolutionary
dynamics without directly altering the task objective.

\subsubsection{Operators and hyperparameters}

Table \ref{tab:operators_and_hyper} summarizes the important operators of our algorithm-discovery framework, and the relevant hyperparameters for contrastive concept tree search, the algorithm proposed in this work.

\begin{table}[h]
\centering
\caption{Summary of operators used in the LLM-assisted algorithm discovery framework, and main hyperparameters $\theta$ of the Contrastive Concept-Tree Search (CCTS) framework. }
\small
\begin{tabular}{p{2cm} p{6.0cm} p{7cm}}
\hline
\textbf{Operator} & \textbf{Description} & \textbf{Implementation} \\
\hline
$\rho$ & Strategy / policy selector & Fixed core kernel (exploration-exploitation mixture) \\
$g$ & Meta context generator & LLM-assisted, via prompting with child and parent code \\
$\pi$ & Parent selector & Fixed core kernel (CCTS, Greedy, Uniform, k-elites) \\
$q$ & Child generator & LLM-assisted, via prompting with parent code \\
$\Phi$ & Feature extractor & LLM-assisted, via prompting with child and concept tree \\
$E$ & Fitness function of task $C$ & Task-dependent, user-defined \\
\hline
\hline
\textbf{Symbol} & \textbf{Description} & \textbf{Default} \\
\hline

\multicolumn{3}{l}{\textit{LLM-related}} \\
\hline
$T_{\mathrm{LLM}}$ 
& LLM sampling temperature
& $0.75$ \\

$\mathcal{L}$ 
&  LLM model used
& \texttt{gemini-2.0-flash} \\

\hline
\multicolumn{3}{l}{\textit{Search dynamics}} \\
\hline
$p_{\mathrm{exploit}}$ 
& Exploitation weight 
& $0.85$ \\

$T$ 
& Iteration per run 
& $25$ \\

\hline
\multicolumn{3}{l}{\textit{New concept suggestion mechanism}} \\

\hline
$\lambda$ 
& Exploration weight in concept proposal 
& $0.05$ \\

$\gamma$ 
& Novelty decay exponent for concept selection 
& $1.0$ \\




\hline
\multicolumn{3}{l}{\textit{Population structure}} \\
\hline
$K$ 
& Number of parallel islands 
& $1$ \\



\hline
\end{tabular}
\label{tab:operators_and_hyper}
\end{table}

\subsection{Learning model description\label{sec:fac_MLE_appendix}}

In this section, we describe the model used to infer the class-conditional probability distribution used by the student in the CCTS framework to approximate the log-likelihood  for a given concept to belong to a high quality or a low-quality algorithm (see Sec. \ref{sec:comparison}). 

\subsubsection{Factorized categorical case}

Let $b = (b_v)_{v \in V}$ denote a binary vector of concept indicators extracted from an algorithm, where
$V$ indexes concepts organized in a rooted tree with parent map $\mathrm{pa}(v)$. Such a tree is constructed by directly querying the LLM for empirical tasks, or through a teacher model for synthetic tasks, which is going to be described in detail in App. \ref{sec:synthetic}.
We consider a parametric family of distributions over $b$ in which features are conditionally independent
given their parent in the hierarchy and satisfy a hard ancestor-consistency constraint.

Specifically, we assign zero probability to any configuration violating
\begin{align}
b_v \le b_{\mathrm{pa}(v)} \qquad \text{for all } v \in V,
\end{align}
so that a concept can be active only if its parent concept is active too.
For valid configurations obeying the condition above, we use a hierarchical Bernoulli model in which
features are conditionally independent, given the (necessary) presence of their parent. In this factorized case, the parameter vector $\eta$ of the generic family
$\{\hat p_\eta(b)\}$ consists of per-node conditional activation probabilities
$\eta = \{\eta_v\}_{v\in V}$, defined by
\begin{align}
\Pr(b_v = 1 \mid b_{\mathrm{pa}(v)} = 1) = \eta_v,
\qquad
\eta_v \in (0,1),
\end{align}
with $\Pr(b_v = 1 \mid b_{\mathrm{pa}(v)} = 0) = 0$ by construction.

Under this parameterization, the induced joint distribution over valid configurations factorizes as
\begin{align}
\hat{p}_\eta(b)
=
\prod_{v \neq \text{root}}
\eta_v^{\, b_v}
(1 - \eta_v)^{\, b_{\mathrm{pa}(v)} - b_v},
\end{align}
and $\hat{p}_\eta(b)=0$ for any configuration violating the hierarchical constraint $b_v \le b_{\mathrm{pa}(v)}$.

To guide parent selection, we adopt a Tree-structured Parzen Estimator (TPE) formulation and estimate two such models with identical structure:
$\hat{p}_{\eta^+}(b) \approx p(b \mid \text{good}_t)$
and
$\hat{p}_{\eta^-}(b) \approx p(b \mid \text{bad}_t)$.
The parameters are learned by cross-entropy / maximum likelihood estimation\footnote{When the target distribution is empirical, minimizing the KL divergence (or equivalently the cross-entropy) between the empirical measure and a parametric model is exactly equivalent to maximum likelihood estimation.} (a detailed derivation is presented in the paragraph below) on the corresponding subsets of the archive,
\begin{align}\label{eq:app_CE_estimates}
    \begin{split}
    \hat{\eta}^{+}_v
    &=
    \frac{\sum_{i \in \text{good}_t} b_{i,v}}
         {\sum_{i \in \text{good}_t} b_{i,\mathrm{pa}(v)}}\,,
    \\
    \hat{\eta}^{-}_v
    &=
    \frac{\sum_{i \in \text{bad}_t} b_{i,v}}
         {\sum_{i \in \text{bad}_t} b_{i,\mathrm{pa}(v)}}\,.
    \end{split}
\end{align}

Guided parent selection is then based on the log-likelihood ratio
\begin{align}
\log w(b)
=
\log \hat{p}_{\eta^+}(b)
-
\log \hat{p}_{\eta^-}(b),
\end{align}
which assigns high weight to algorithms whose semantic features are enriched in high-performing solutions relative to low-performing ones. 

\paragraph{Factorized hierarchical Bernoulli model}

For the sake of completeness, we derive here the cross-entropy estimates given in Eq. \eqref{eq:app_CE_estimates}. For the factorized hierarchical Bernoulli model, the log-likelihood of a single feature vector
$b = (b_v)_{v\in V}$ is
\begin{align}
\log \hat p_\eta(b)
=
\sum_{v \neq \text{root}}
\Big[
b_v \log \eta_v
+
\bigl(b_{\mathrm{pa}(v)} - b_v\bigr)\log(1-\eta_v)
\Big],
\end{align}
with the convention that $\hat p_\eta(b)=0$ for configurations violating
$b_v \le b_{\mathrm{pa}(v)}$. Given a subset of the archive (either $\mathrm{good}$ or $\mathrm{bad}$), maximum likelihood estimation
amounts to maximizing the sum of the log-likelihood above over all the samples in the subset.
Because the likelihood factorizes across nodes, the optimization problem decouples over the different vector entries $v$. We then define, for a fixed node $v$, define
\begin{equation}\label{eq:app_ABpm}
A_v^\pm \equiv \sum_{i \in \mathrm{good/bad}} b_{i,v},
\qquad
B_v^\pm \equiv \sum_{i \in \mathrm{good/bad}} b_{i,\mathrm{pa}(v)},
\end{equation}
where the sums run over the indices $i$ in the selected subset (either $i \in \mathrm{good}$ or $i \in \mathrm{bad}$).
The contribution of node $v$ to the total log-likelihood in either of the two cases, $\ell_v^{\pm}(\eta_v)$, is then given by
\begin{equation}
\ell^\pm_v(\eta_v)
=
A_v^\pm \log \eta_v
+
(B_v^\pm - A_v^\pm)\log(1-\eta_v).
\end{equation}

Setting $\partial \ell^\pm_v / \partial \eta_v \rvert_{\eta=\eta_v^\pm}= 0$ yields the closed-form estimator
\begin{equation}\label{eq:app_theta_mle}
    \hat{\eta}_v^\pm = \frac{A_v^\pm}{B_v^\pm}\,,
\end{equation}
which corresponds to the empirical conditional probability
$\Pr(b_v = 1 \mid b_{\mathrm{pa}(v)} = 1)$ estimated from the selected subset.

\subsection{New concept exploration \label{sec:concept_exploration}}

The guided parent-selection mechanism described above biases sampling toward archive elements whose existing semantic features are enriched among high-performing algorithms. However, this alone does not actively encourage the exploration of new or rarely tried concepts. We augment CCTS with a lightweight concept-level exploration mechanism that operates directly on the concept tree, and that we describe hereafter.

\paragraph{Beta priors and nonzero support.}
To ensure that newly discovered concepts are never assigned zero probability, we place Beta priors on the conditional activation parameters of the hierarchical Bernoulli model. Concretely, for each concept $v$, rather than using the maximum likelihood estimators given by Eq. \eqref{eq:app_theta_mle}, we estimate the parameters $\eta_v^{\pm}$ through smoothed counts,
\begin{equation}\label{eq:app_tilde_etav_pm}
\tilde{\eta}_v^{\pm}
=
\frac{A_v^{\pm} + \alpha_0}{B_v^{\pm} + \alpha_0 + \beta_0},
\end{equation}
where $A_v^{\pm}$ and $B_v^{\pm}$ denote the empirical counts in the $\mathrm{good}$ or $\mathrm{bad}$ subsets of the archive, and are given by Eq. \eqref{eq:app_ABpm}, and $\alpha_0$ and $\beta_0$ are smoothening parameters ($\alpha_0 = 0.1$ and $\beta_0 = 10.0$). This guarantees nonzero probability mass for all concepts, including those that have not yet been observed, and stabilizes likelihood-ratio estimates for guided selection.

\paragraph{Count-based novelty bias.}
In addition to exploitation driven by the likelihood-ratio score, we introduce an explicit bias toward concepts that have been tried fewer times. For each concept $v$, we maintain a counter $n_v$ representing the number of times that concept has been targeted during child generation. We define a novelty weight
\begin{equation}
\mathrm{novel}(v) = (n_v + 1)^{-\gamma},
\end{equation}
where $\gamma > 0$ controls the strength of the bias. Concepts that have never been tried ($n_v=0$) receive maximal novelty weight, while repeatedly attempted concepts are progressively downweighted. This mechanism naturally supports both first-time exploration and repeated trials when the effect of
a concept is uncertain.

\paragraph{Mixture-based concept selection.}
The Beta priors and the count-based novelty bias both contribute to the probability of $p(v)$ of proposing a semantic edit for a new child during the evolution process. Candidate concepts are sampled from a mixture of an exploitation distribution based on the learned contrastive model and an exploration distribution based on novelty, namely,
\begin{equation}
p(v) \equiv 
(1-\lambda)\,
\frac{\exp(\Delta_v)}{\sum_{u}\exp(\Delta_u)}
\;+\;
\lambda\,
\frac{\mathrm{novel}(v)}{\sum_{u}\mathrm{novel}(u)},
\qquad
\Delta_v = \log \tilde{\eta}_v^{+} - \log \tilde{\eta}_v^{-},
\end{equation}
where $\lambda\in[0,1]$ is a parameter that controls the trade-off between exploitation, given by consecutive, incremental, lineage based mutations, and exploration, which biases the generation process toward unprecedentedly used concepts. 

\paragraph{Prompt-level realization.}
Once a concept $v$ is selected, it is injected into the LLM prompt as a semantic directive for child generation. Specifically, the adaptive prompt is augmented as
\begin{align}
P_t = P_0(C) \;\oplus\; \mathrm{Ctx}_t,
\end{align}
\noindent where $\mathrm{Ctx}_t = \texttt{``Try to incorporate concept $v$''}$.

\subsection{Synthetic environment \label{sec:synthetic}}
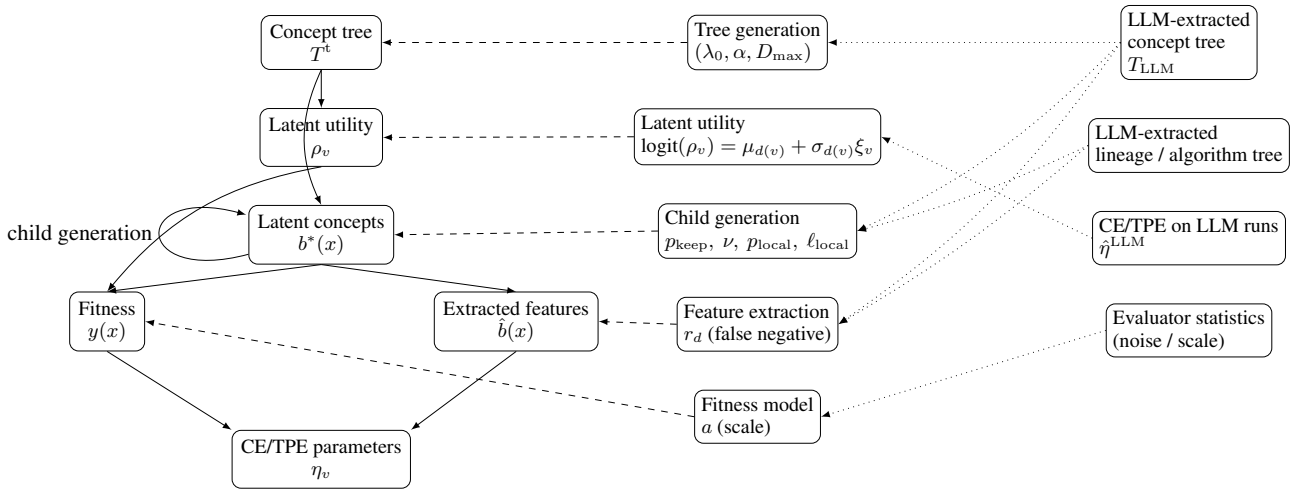
\begin{figure}[h]
\centering
\resizebox{\textwidth}{!}{%
\begin{tikzpicture}[
    node distance=0.6cm,
    main/.style={draw, rounded corners, align=center, font=\small, inner sep=4pt},
    param/.style={draw, rounded corners, align=left, font=\small, inner sep=3pt},
    data/.style={draw, rounded corners, align=left, font=\small, inner sep=3pt},
    >=latex
]

\node[main] (tree) {Concept tree\\ $T^{\mathrm t}$};

\node[main, below=of tree] (rho) {Latent utility\\ $\rho_v$};

\node[main, below=of rho] (bstar) {Latent concepts\\ $b^*(x)$};

\node[main, below left=of bstar, xshift=-1.2cm] (fitness) {Fitness\\ $y(x)$};
\node[main, below right=of bstar, xshift=0.2cm] (bhat) {Extracted features\\ $\hat b(x)$};

\node[main, below=of bstar, yshift=-2.0cm] (eta) {CE/TPE parameters\\ $\eta_v$};

\draw[->] (tree.south) -- (rho.north);
\draw[->] (tree.south) to[bend right=25] (bstar.north);

\draw[->] (rho.south) to[bend right=20] (fitness.north);
\draw[->] (bstar.south) -- (fitness.north);

\draw[->] (bstar.south) -- (bhat.north);

\draw[->] (fitness.south) -- (eta.north west);
\draw[->] (bhat.south) -- (eta.north east);

\draw[->] (bstar) edge[loop left] node {child generation} ();

\node[param, right=4.8cm of tree] (p_tree)
{Tree generation\\
$(\lambda_0,\alpha,D_{\max})$};

\node[param, below=of p_tree] (p_theta)
{Latent utility\\
$\text{logit}(\rho_v)=\mu_{d(v)}+\sigma_{d(v)}\xi_v$};

\node[param, below=of p_theta] (p_child)
{Child generation\\
$p_{\mathrm{keep}},\ \nu,\ p_{\mathrm{local}},\ \ell_{\mathrm{local}}$};

\node[param, below=of p_child] (p_feat)
{Feature extraction\\
$r_d$ (false negative)};

\node[param, below=of p_feat] (p_fit)
{Fitness model\\
$a$ (scale)};

\draw[->, dashed] (p_tree.west) -- (tree.east);
\draw[->, dashed] (p_theta.west) -- (rho.east);
\draw[->, dashed] (p_child.west) -- (bstar.east);
\draw[->, dashed] (p_feat.west) -- (bhat.east);
\draw[->, dashed] (p_fit.west) -- (fitness.east);

\node[data, right=4.6cm of p_tree] (d_tree)
{LLM-extracted\\ concept tree\\ $T_{\mathrm{LLM}}$};

\node[data, below=of d_tree] (d_lineage)
{LLM-extracted\\ lineage / algorithm tree};

\node[data, below=of d_lineage] (d_eta)
{CE/TPE on LLM runs\\ $\hat\eta^{\mathrm{LLM}}$};

\node[data, below=of d_eta] (d_eval)
{Evaluator statistics\\ (noise / scale)};

\draw[->, dotted] (d_tree.west) -- (p_tree.east);
\draw[->, dotted] (d_eta.west) -- (p_theta.east);

\draw[->, dotted] (d_tree.west) to[bend left=10] (p_child.east);
\draw[->, dotted] (d_lineage.west) to[bend left=5] (p_child.east);

\draw[->, dotted] (d_tree.west) to[bend left=10] (p_feat.east);
\draw[->, dotted] (d_lineage.west) to[bend left=5] (p_feat.east);

\draw[->, dotted] (d_eval.west) -- (p_fit.east);

\end{tikzpicture}%
}
\caption{Schematic description of the synthetic environment. Left: generative teacher model. Center: hyperparameters controlling each component. Right: empirical statistics extracted from LLM runs used to fit or calibrate the hyperparameters. The solid arrow connect an element of the synthetic environment with what is constructed starting from it. The dashed lines connect a set hyperparameters with the object they define. The dotted lines connect an empirical  statistics from real tasks to the synthetic hyperparameters that fit them.}
\label{fig:artificial_data}
\end{figure}

In this Section, we describe the synthetic environment used to rationalize our empirical findings. A schematic summary of the key elements composing the synthetic environment is shown in Fig. \ref{fig:artificial_data}, while a table summarizing the notation pertaining to the synthetic environment is reported in Table \ref{tab:synthetic_parameters} and are precisely characterized in what follows.

\begin{table}[t]
\caption{Parameters of the synthetic algorithm-discovery environment.}
\centering
\small
\begin{tabular}{p{2.8cm} p{2.2cm} p{8.5cm}}
\hline
\textbf{Symbol} & \textbf{Value} & \textbf{Description} \\
\hline

\multicolumn{3}{l}{\textit{Concept tree generation (branching process)}} \\
\hline
$\lambda_0$ & $5.0$ 
& Expected branching factor at the root of the concept tree. \\

$\alpha$ & $0.9$ 
& Exponential decay rate of branching with depth. \\

$D_{\max}$ & $10$ 
& Maximum depth of the concept tree. \\

$N_{\max}$ & $25$ 
& Maximum number of concepts (used in Fig.~\ref{fig:exploit_vs_lambda}). \\

\hline
\multicolumn{3}{l}{\textit{Latent mutation operator (synthetic child generation)}} \\
\hline
$p_{\mathrm{keep}}$ & $0.75$ 
& Probability of inheriting an active concept from the parent. \\

$\nu$ & $1.5$ 
& Mean number of new concepts introduced per mutation. \\

$p_{\mathrm{local}}$ & $0.65$ 
& Probability of sampling new concepts locally in the tree. \\

\hline
\multicolumn{3}{l}{\textit{Ground-truth concept utility (teacher model)}} \\
\hline
$\mu_0$ & $0.0$ 
& Mean logit utility of depth-$0$ concepts. \\

$\sigma_0$ & $1.0$ 
& Standard deviation of logit utility at depth $0$. \\

\hline
\multicolumn{3}{l}{\textit{Noise and observation model}} \\
\hline
$\sigma_y$ & $0.1$ 
& Standard deviation of fitness noise. \\

\hline
\end{tabular}
\label{tab:synthetic_parameters}
\end{table}

\subsubsection{Goal}

We introduce a synthetic algorithm-discovery environment that replaces the LLM with controlled probabilistic operators. The purpose is to (i) isolate the structural mechanisms underlying Contrastive Concept-Tree Search, (ii) identify optimal operating regimes (e.g., locality, noise, concept granularity), and (iii) explain empirical scaling behavior observed with real LLMs.

The synthetic environment mirrors the real system at the level of semantics rather than language: algorithms are represented by latent semantic concepts organized in a hierarchy; fitness depends on which concepts are present within a given algorithm; child generation performs local semantic edits; and the feature extraction process is noisy. The guided selection mechanism operates on extracted features exactly as in the real
system, by fitting paired feature models over good and bad subsets and using their likelihood ratio to reweight parent selection.

\subsubsection{Model\label{sec:analysis_model}}

In this Section, we detail how the different components of the synthetic setting are constructed. These components are: a teacher structure containing the possible concepts and their tree-structured, semantic relations; a semantic utility function that associates to each concept a numerical value representing its utility; a "teacher truth", i.e. a set of concepts through which an algorithm is defined; a synthetic mutation process that generates a child from a given parent, and that replaces the LLM as a mutation operator; a linear teacher evaluator that assigns to a given algorithm a score, or fitness, and a synthetic learner that extract an imperfect set of concepts from a given algorithm. 

\paragraph{Concept tree generation (teacher structure)}

An artificial latent (teacher) concept tree is generated by a truncated, depth-inhomogeneous Galton--Watson branching process \cite{Draief_Massoulié_2009}. Starting from the root node $b^{\mathrm t}_0$, each node $b$ at depth $d(b)$ generates $K_b$ offspring, where $K_b$ is drawn from a Poisson distribution with depth dependent rate $\lambda_d$, namely, 
\begin{align}
K_b \mid d(b)=d \sim \mathrm{Poisson}(\lambda_d),
\qquad
\lambda_d = \lambda_0 \exp(-\alpha d),
\end{align}
The base branching ratio $\lambda_0 > 0$ controls the branching process at the root and $\alpha \ge 0$ controls the decay of the mean number of offspring with the depth reached. The process is truncated at a maximum depth $D_{\max}$ by setting $K_b = 0$ for all $d(b)=D_{\max}$. Here, the parameter $\lambda_d$ is the mean (and variance) of the Poisson distribution, so that $\mathbb{E}\!\left[K_b \mid d(b)=d\right] = \lambda_d$ and $\mathrm{Var}\!\left(K_b \mid d(b)=d\right) = \lambda_d$. The hyperparameters $(\lambda_0,\alpha,D_{\max})$ are fitted to LLM-extracted concept trees by approximately matching its empirical mean branching factor as a function of depth. The resulting set of nodes in the tree, denoted by $V$, constitutes the set of existing concepts in the synthetic task.

\paragraph{Teacher concept values (semantic utility)}

Each concept $v\in V$ is assigned a latent utility parameter $\rho_v \in (0,1)$, which controls how strongly the presence of that concept contributes to the quality of the algorithm. We model the utility of a concept using an underlying Gaussian distribution on the logit scale (i.e., $\rho_v$ is the sigmoid of a Gaussian random variable):
\begin{align}
\text{logit}(\rho_v) \equiv \log\!\bigl(\rho_v/(1-\rho_v)\bigr) = \mu_{d(v)} + \sigma_{d(v)}\,\xi_v,
\qquad
\xi_v \sim \mathcal N(0,1),
\end{align}

\noindent where $\mu_d$ denotes average utility of concepts at depth $d$, and $\sigma_d$ denotes the heterogeneity of utility at that depth. $\xi_v$ is a Gaussian noise with zero mean and unit correlation. For simplicity, we assume that the realizations of the noise for different concept values are independent from each other. The logit parametrization ensures that the latent utility parameters $\rho_v$ remain in $(0,1)$ while allowing their distribution to be modeled with an unconstrained Gaussian prior on the logit scale.

\paragraph{Latent algorithm representation (teacher truth)}

Each algorithm $x$ is associated with a latent concept configuration $b^*(x) \in \{0,1\}^{|V|}$, subject to a hierarchical closure based on the teacher structure: for every non-root node $v\in V$,
\begin{align}
b^*_v(x)=1 \;\Rightarrow\; b^*_{\mathrm{pa}(v)}(x)=1.
\end{align}

The teacher truth $b^*(x)$ represents which semantic concepts the algorithm truly uses and is not observed directly by the learner. In this way, we model the possible existence of a mismatch between the LLM's representation of a given algorithm and what the algorithm actually does to accomplish a given task.

\paragraph{Child generation operator (synthetic mutation)}

Child generation operates directly on latent concepts $b^*$ and induces local semantic search. Given a parent with latent concepts $b^*_{\text{parent}}$, we first retain each active concept  with probability $p_{\mathrm{keep}}$ (this step models a possible "inheritance" mechanism). We then sample $m\sim \mathrm{Poisson}(\nu)$ new concepts with probability $p_{\mathrm{local}}$. These concepts are chosen via a short random walk on the tree, and represent forms of gradual innovation. Alternatively we sample uniformly from the tree (radical innovation), with probability $1-p_\mathrm{local}$. Lastly, we activate all ancestors of any newly activated concept, thus accommodating the child algorithm within the teacher structure. 

This evolution operator induces a notion locality and smoothness within the the search landscape. The gradual innovation probability $p_{\mathrm{local}}$ and the mean of the Poisson process $\nu$ are selected so that synthetic model exhibits search behavior qualitatively similar to that seen in the LLM-based experiments.

\paragraph{Fitness generation (linear teacher evaluator)}

The fitness of an algorithm $x$ is generated as a noisy perturbation of the latent values $w_v$ of the latent concepts $b^*(x)$ represented by $x$, namely,
\begin{align}
y(x) = \sum_{v\in b^*(x)} w_v + \varepsilon,
\qquad
\varepsilon \sim \mathcal N(0,\sigma_y^2),
\end{align}
with
\begin{align}
w_v = \bigl(\text{logit}(\rho_v)-\bar\mu_{d(v)}\bigr),
\qquad
\bar\mu_d = \mathbb E_{v:d(v)=d}\!\left[\text{logit}(\rho_v)\right].
\end{align}

The limitation of this simple model is that the fitness $y(x)$ is assumed to be a linear function of the concept utility. A more refined model would account for dependencies between concepts, but this is out of the scope of this work.

\paragraph{Feature extraction (synthetic learner noise)}

We denote by $\hat b(x)$ the concepts that the learner attributes to  a given algorithm. We assume that the learner observes a noisy version of the latent concepts:
\begin{align}
\hat b(x) = \Phi\bigl(b^*(x)\bigr).
\end{align}

We construct the observed concepts as follows: starting from $b^*(x)$, any concept within it can be dropped with probability $1-r_d$. Hierarchical closure is then enforced on the resulting vector, yielding $\hat b$. This prescription models an imperfect concept extraction from the LLM. The imperfection modeled yields false negatives (a concept belonging to the algorithm is missed by the LLM).  Although not considered here, it is also possible to model the addition of spurious concepts, or false positives.

\subsubsection{Comparison with the LLM-generated data}\label{sec:comparison}

\paragraph{Learning mechanism (CE / TPE)}

The learner applies a contrastive update in the style of a Tree Parzen Estimator on the extracted features $\hat b$. More precisely,
at iteration $t$, the archive is partitioned into good and bad sets using a threshold $\tau_t$
(e.g., top-$\rho$ quantile):
\begin{align}
\text{good}_t = \{\, i : y_i \ge \tau_t \,\}, \qquad
\text{bad}_t = \{\, i : y_i < \tau_t \,\}.
\end{align}
Two class-conditional feature models are then fitted to the conditional probabilities that a good or bad algorithm as a given concept $b$ as active,
\begin{align}
\hat p_{\eta^+}(\hat b) \approx p(\hat b \mid \text{good}_t),
\qquad
\hat p_{\eta^-}(\hat b) \approx p(\hat b \mid \text{bad}_t),
\end{align}
given that the hierarchical closure induced by the concept tree model (
Section~\ref{sec:method}) is respected. Guided parent selection according to the contrastive concept search prediction uses then the log-likelihood ratio  $\log w(\hat b)$ as a discriminative score to select the parent during the synthetic mutation process, namely,
\begin{align}
\log w(\hat b)
=
\log \hat p_{\eta^+}(\hat b) - \log \hat p_{\eta^-}(\hat b),
\qquad
\pi(x_i \mid \mathcal{A}_t) \propto \exp(\log w(\hat b_i))\,,
\end{align}
with $\pi(x_i|\mathcal{A}_t)$ the probability to select algorithm $x_i$ as a parent for the synthetic evolution process.  In summary, $\rho_v$ is the ground-truth latent utility driving the teacher fitness, while the learned parameters $\eta^+,\eta^-$ (and the induced ratio score $\log w$) are emergent statistics that drive selection toward more successful concepts, and that rely on an imperfect, noisy learner. 

\paragraph{Comparison between teacher and student}
\label{sec:teacher_student_comparison}

The synthetic environment allows us to explicitly compare the latent ground-truth structure used to generate data, i.e. the teacher, with the statistics learned by the guided search procedure, the student. Importantly, this comparison must respect the fact that the learner does not attempt to recover the teacher parameters directly, but instead learns discriminative feature statistics induced by selection dynamics and noisy observations. The teacher is characterized by a fixed set of latent utility parameters
$\rho = \{\rho_v\}_{v\in V}$, which determine the fitness of a given algorithm through the data-generating process (Section~\ref{sec:analysis_model}). These parameters define how individual concepts contribute to fitness in expectation, but they do not depend on the archive state nor on any partition between arbitrarily defined "good" and "bad" concepts. By contrast, the learner explicitly introduces a partition in algorithmic space, by fitting two class-conditional feature models,
$\hat p_{\eta^+}(b) \approx p(b \mid \text{good})$ and
$\hat p_{\eta^-}(b) \approx p(b \mid \text{bad})$,
and uses their likelihood ratio to guide parent selection.
The quantity that actually drives selection is therefore the
discriminative score
\begin{align}
\log w(b)
=
\log \hat p_{\eta^+}(b) - \log \hat p_{\eta^-}(b),
\end{align}
which measures how strongly a feature configuration is enriched among high-performing algorithms relative to low-performing ones. Thus, to compare teacher and student on a common footing, we derive a corresponding
teacher-side discriminative signal by applying the same partition among "good" and "bad" algorithms in the synthetic archive and computing, for each concept $v$, the conditional log-odds ratio
\begin{align}
\log w_v^{\text{(teacher)}}
=
\log
\frac{
\Pr(b_v^*=1 \mid b_{\mathrm{pa}(v)}^*=1,\; y \ge \tau)
}{
\Pr(b_v^*=0 \mid b_{\mathrm{pa}(v)}^*=1,\; y \ge \tau)
}
-
\log
\frac{
\Pr(b_v^*=1 \mid b_{\mathrm{pa}(v)}^*=1,\; y < \tau)
}{
\Pr(b_v^*=0 \mid b_{\mathrm{pa}(v)}^*=1,\; y < \tau)
}.
\end{align}
This quantity is not a parameter of the teacher model, but an empirical statistic implied by the teacher’s data-generating process under the same selection criterion used by the learner. The student-side analogue is given by the learned parameters,
\begin{align}
\log w_v^{\text{(student)}}
=
\log \tilde{\eta}_v^+ - \log \tilde{\eta}_v^-\,,
\end{align}
where $\tilde \eta^{\pm}_v$ are the maximum likelihood estimators with smoothed counts given by Eq. \eqref{eq:app_tilde_etav_pm}. The quantity $\log w_v^\text{student}$ corresponds to the per-concept contribution to the log-likelihood ratio used in parent selection.

Comparing $\log w_v^{\text{(teacher)}}$ and $\log w_v^{\text{(student)}}$ therefore isolates how faithfully the guided search mechanism recovers the directional improvement signal encoded in the teacher, rather than attempting to match absolute concept utility. Discrepancies between the two log-likelihoods reflect the combined effects of noisy feature extraction, finite sample size, hierarchical dependencies, and selection-induced bias, and provide a principled way to analyze when and why concept-guided search succeeds or fails in recovering meaningful semantic structure.

In the experiments reported here, the teacher-side discriminative statistic is computed using a long-time population reference obtained by Monte Carlo sampling of the synthetic world (implemented as an MCMC over latent algorithms, here identified by a set of concepts consistent with the teacher's ground truth tree), which defines a fixed partition among good and bad algorithms independent of the learner; this stationary reference allows the student–teacher discrepancy to be tracked consistently over training iterations.

\subsection{Additional experimental results}

\subsubsection{Interpretability \label{sec:interpretability}}

\begin{figure}[htbp]
    \centering
    \includegraphics[width=1.0\textwidth]{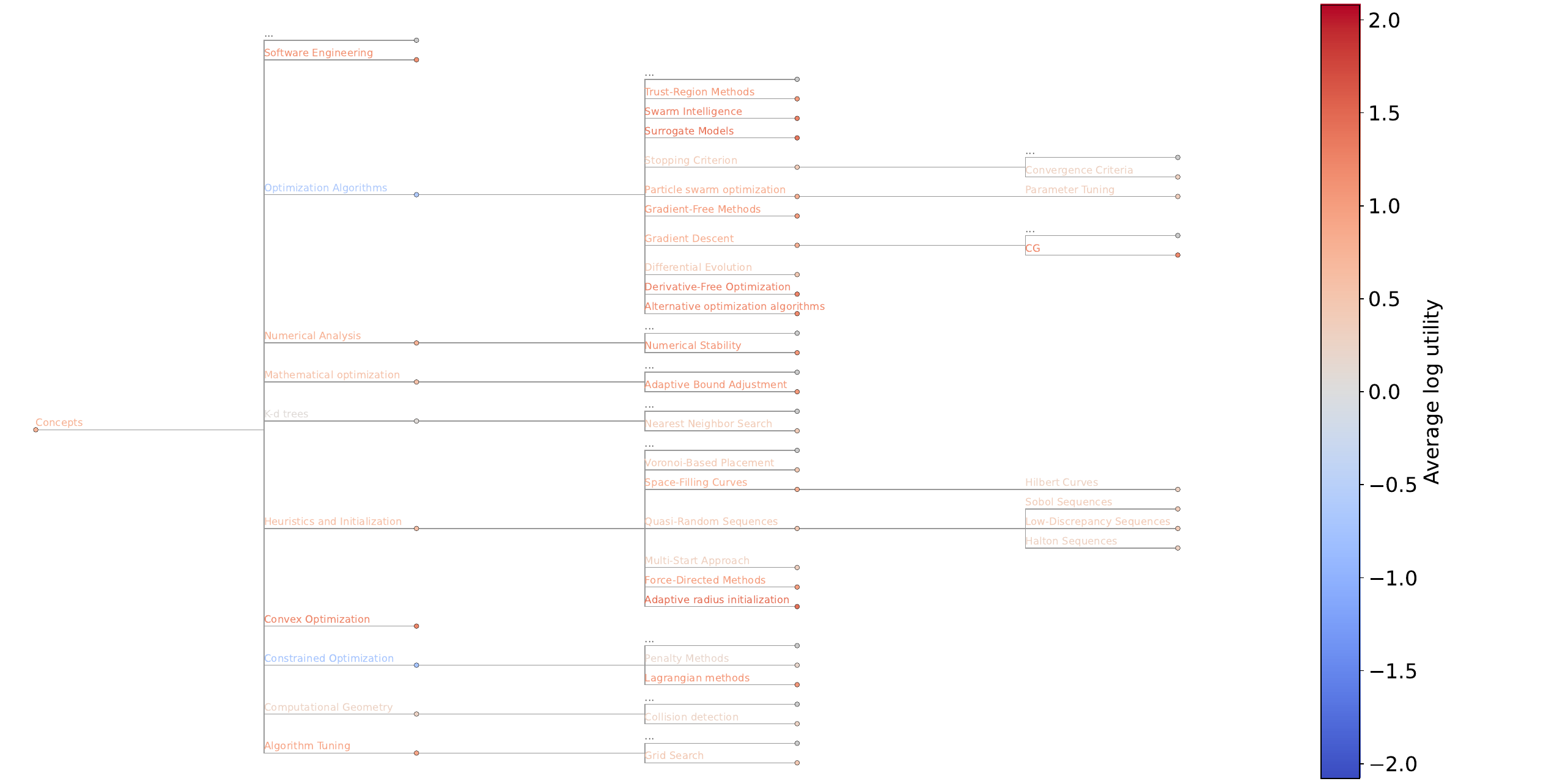}
    \caption{Part of the concept tree extracted by the LLM of the circle packing task. Tree nodes corresponding to the top and bottom 20 inferred concept utilities are shown, with warm and cool colors indicating useful and non-useful concepts, respectively. Utility of concepts are averaged over 60 runs.}
    \label{fig:concept_tree_dendrogram_LLM}
\end{figure}

\begin{figure}[htbp]
    \centering
    \includegraphics[width=0.8\textwidth]{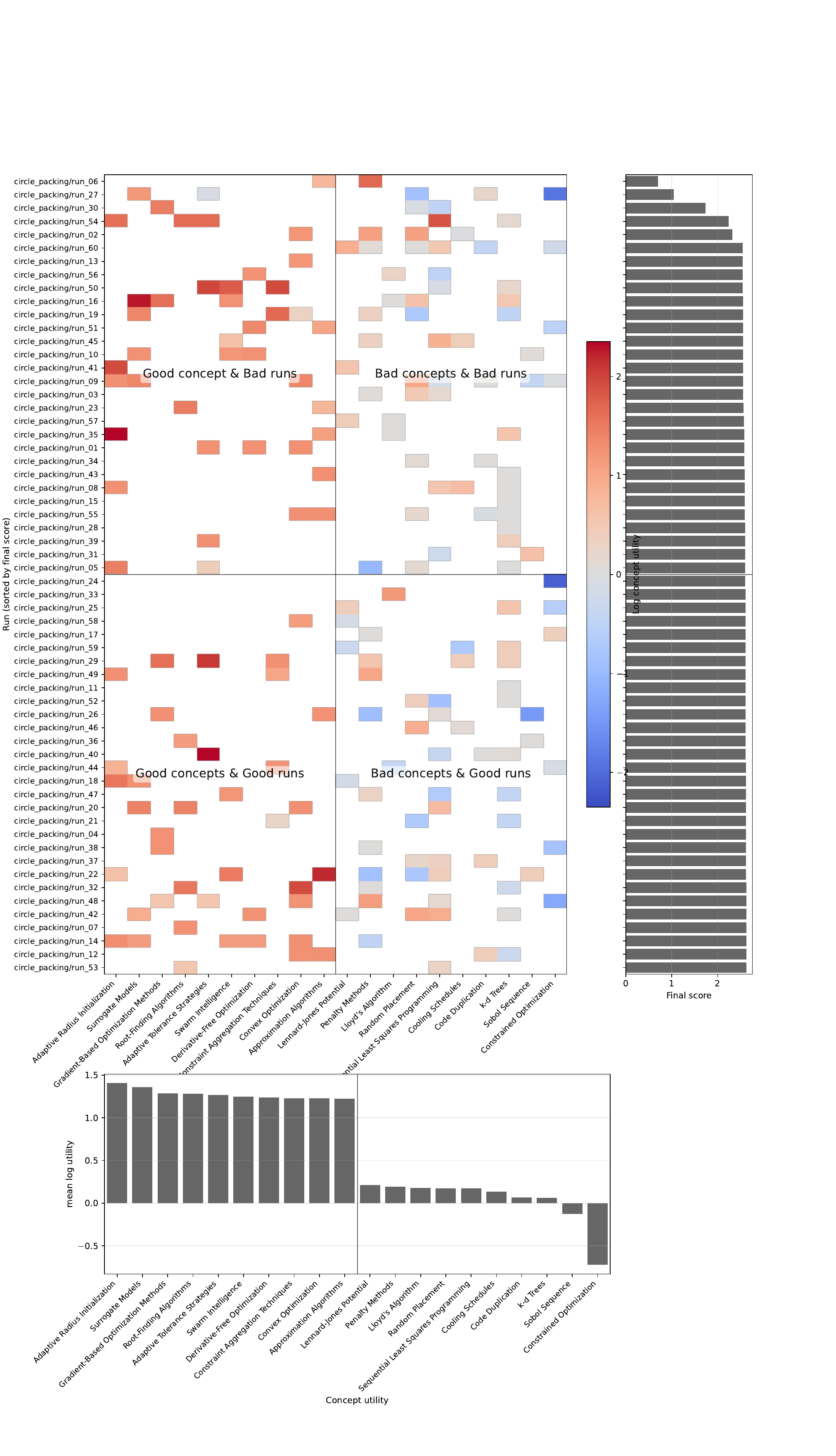}
    \caption{Algorithm score and concept utility scatter plot for the circle packing task.}
    \label{fig:log_lift_summary_circle}
\end{figure}

\begin{figure}[htbp]
    \centering
    \includegraphics[width=1.0\textwidth]{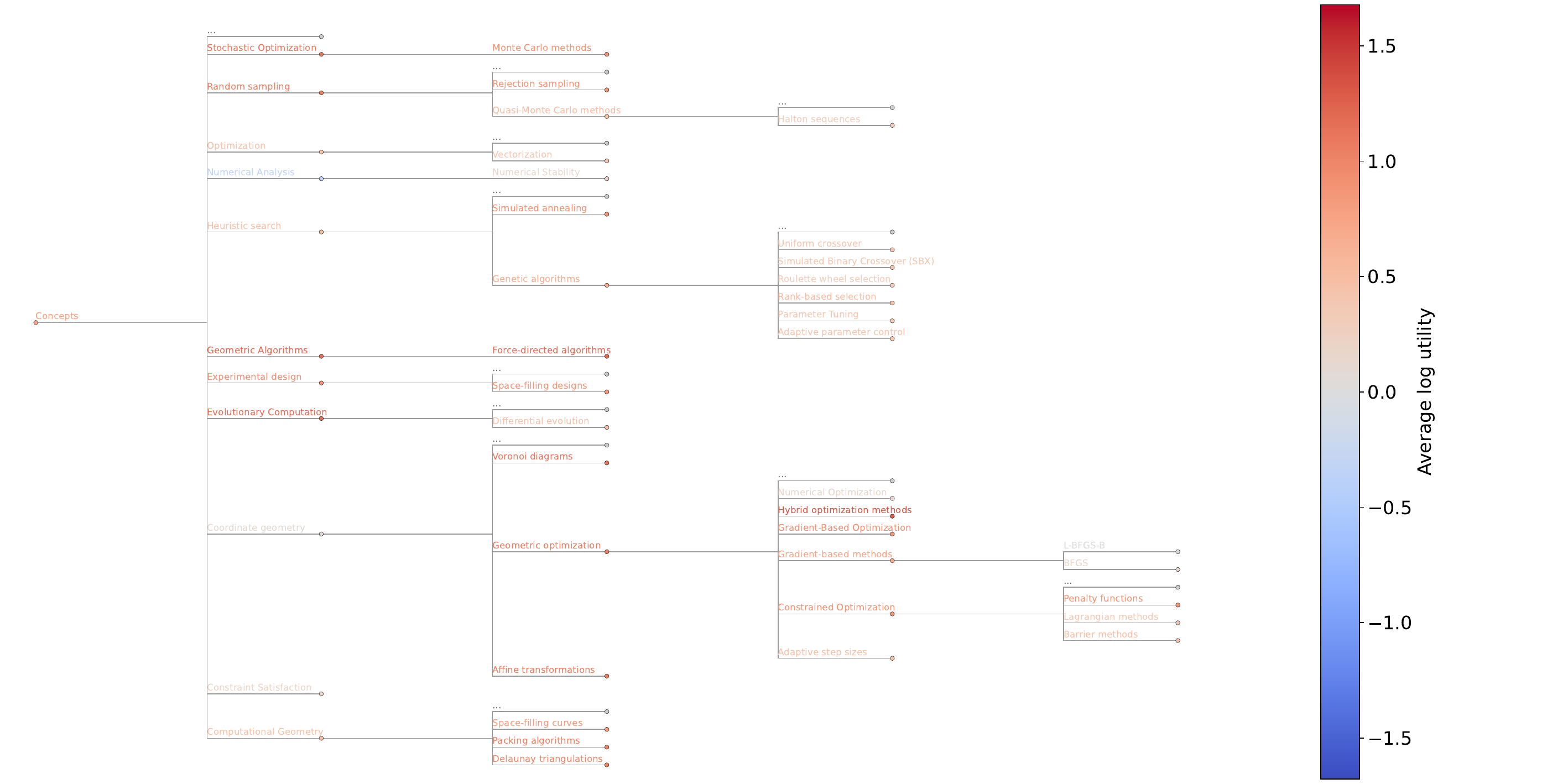}
    \caption{Part of the concept tree extracted by the LLM of the Heilbronn’s Triangle task. Tree nodes corresponding to the top and bottom 20 inferred concept utilities are shown, with warm and cool colors indicating useful and non-useful concepts, respectively. Utility of concepts are averaged over 60 runs.}
    \label{fig:concept_tree_dendrogram_LLM_heilbronn}
\end{figure}

\begin{figure}[htbp]
    \centering
    \includegraphics[width=0.8\textwidth]{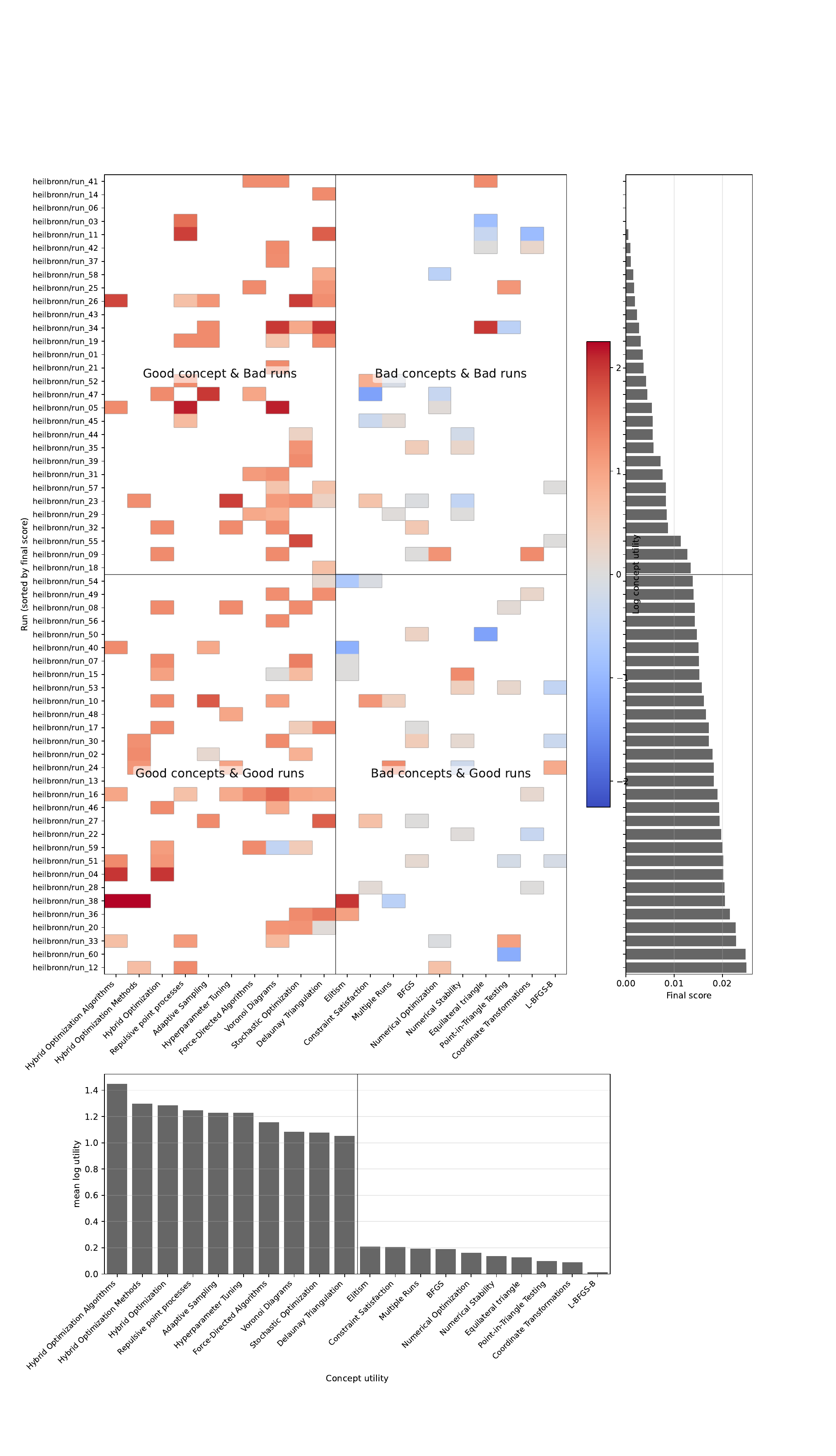}
    \caption{Algorithm score and concept utility scatter plot for the Heilbronn’s Triangle task.}
    \label{fig:log_lift_summary_heilbronn}
\end{figure}

To assess the interpretability of the contrastive concept-tree search, we visualize both the semantic concept trees extracted by the LLM during search (see Figs. \ref{fig:concept_tree_dendrogram_LLM}- \ref{fig:concept_tree_dendrogram_LLM_heilbronn}) and the corresponding concept utility statistics inferred by CCTS (see Figs. \ref{fig:log_lift_summary_circle}- \ref{fig:log_lift_summary_heilbronn}). The figures below illustrate how task-specific concept hierarchies emerge and how individual concepts are associated with algorithm performance.

Figure~\ref{fig:log_lift_summary_circle} visualizes the relationship between algorithm performance and learned concept utility. Runs are ranked by final score along the $y$-axis, while concepts are ranked along the $x$-axis by their learned log utility $\log(w)$, averaged across all runs (showing the top-10 and bottom-10 concepts). Each rectangle describes the log utility learned for a given concept in a given run; red regions denote high-utility (“Good”) concepts and blue points denote low-utility (“Bad”) concepts. Empty entries indicate that the corresponding concept was not explored during that run. Note that the Figure~\ref{fig:log_lift_summary_quadrant_log_lift} is obtained by averaging, across runs, the per-run learned log concept values within the four quadrants defined in Fig.~\ref{fig:log_lift_summary_circle}.


\clearpage

\subsubsection{Growth rate of the number of concepts}

\begin{figure}[t]
    \centering
    \begin{subfigure}[t]{0.49\textwidth}
        \centering
        \caption{Real task}
        \includegraphics[width=\textwidth]{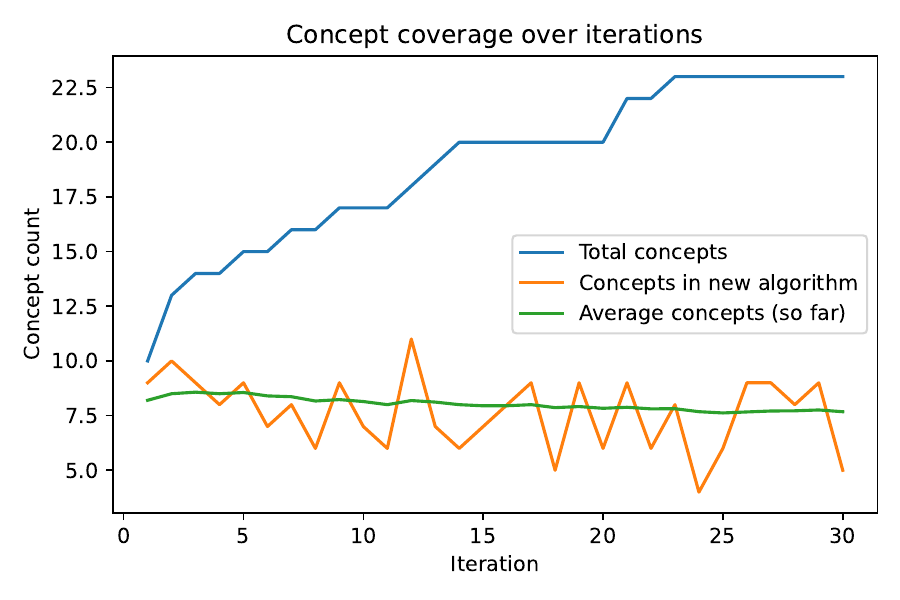}
    \end{subfigure}\hfill
    \begin{subfigure}[t]{0.49\textwidth}
        \centering
        \caption{Synthetic task}
        \includegraphics[width=\textwidth]{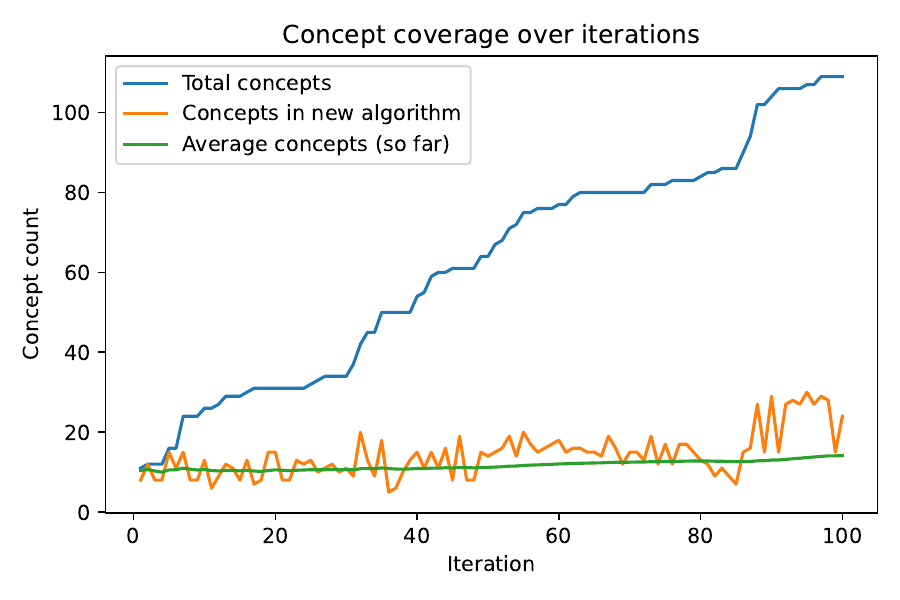}
    \end{subfigure}
    \caption{Number of concepts discovered vs. iteration. The total number of concepts is the cardinality of all concepts appearing in the sampled algorithms at the current iteration.
The number of concepts in the new algorithm refers to the number of concepts in the child algorithm at the current iteration.
The average number of concepts so far is the mean number of concepts across all discovered algorithms.}
    \label{fig:concept_growth}
\end{figure}

The growth of the number of concepts discovered vs. iteration is shown in Fig. \ref{fig:concept_growth} for the circle packing task in the empirical setting and the synthetic task. The curves are qualitatively similar in both the empirical and synthetic settings,
with the total number of discovered concept increasing with the number of iterations. Interestingly, this increase proceeds in a step-like fashion, where time intervals where the total number of concepts remains roughly constant alternate with steep increases in the concept counts. 
\subsubsection{Exploitation/exploration ratio \label{sec:exploit}}

\begin{figure*}[b]
    \centering
    \includegraphics[width=\textwidth]{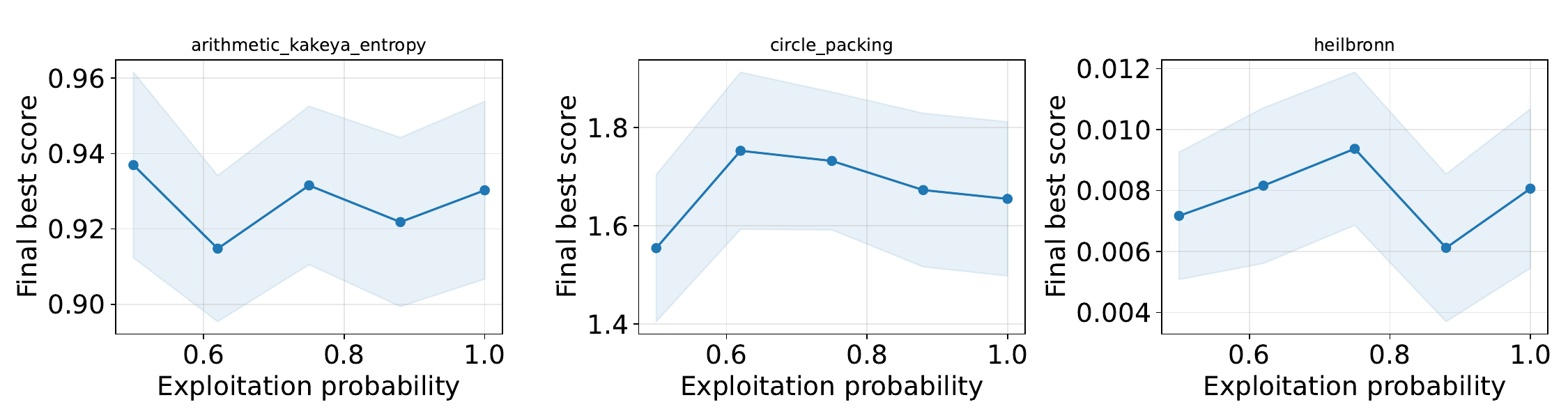}
    \caption{Optimal search strategy of CCTS. We show the normalized score as a function of the exploitation probability $p_{\text{exploit}}$ for the Arithmetic Kakeya conjecture, circle packing, Heilbronn’s triangle problem task after 40 iterations and averaged over 30 runs using gpt-o4-mini. All the curves display a nonmonotonous trend, with a maximum at a value of the exploitation proability that depends on the task.}
    \label{fig:parameter_sweep}
\end{figure*}

In Fig. \ref{fig:parameter_sweep}, we vary the exploitation probability $p_{\text{exploit}}$ while holding all other settings to the main defaults (see Table \ref{tab:operators_and_hyper}), then report the mean final best score across several runs for each value. The experiment is performed using gpt-o4-mini. The curve generically exhibit a nonmonotonic trend, with the optimal exploitation probability varying from task to task. Note that simulations using the synthetic data suggest that the optimal exploitation/exploration strategy depends on the structure, such as the concept branching ratio. of the underlying concept tree (see Fig. \ref{fig:exploit_vs_lambda}).

\subsubsection{Choice of LLM model\label{sec:choiceLLM}}

\begin{figure}[h]
    \centering
    \includegraphics[width=0.6\columnwidth]{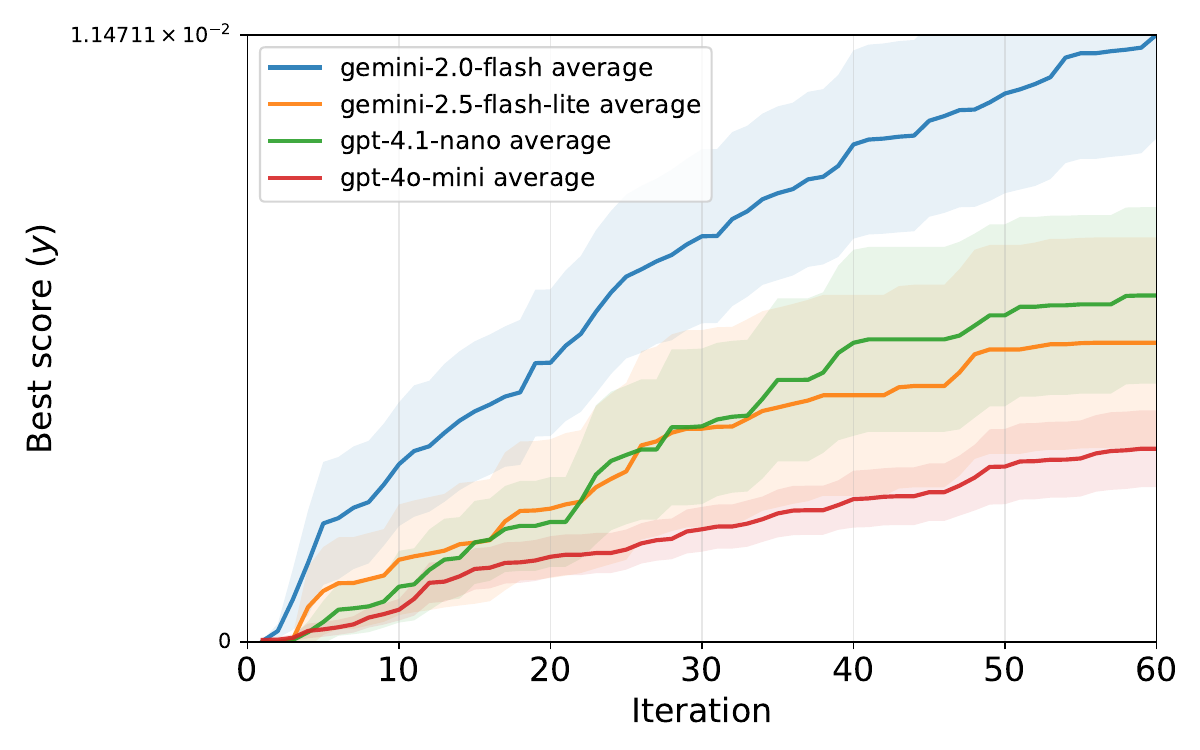}
    \caption{Comparison of LLM models for the Heilbronn Triangle Problem using CCTS, averaged over 60 runs. The shaded area shows the 95\% confidence interval.}
    \label{fig:llm_averages}
\end{figure}

Note that the choice of the underlying LLM model, used in this work as a black-box optimizer, has a significant influence on task performance. We performed experiments with several LLMs exhibiting different reasoning capabilities, showing that more advanced models consistently achieve better performance, as illustrated in Fig.~\ref{fig:llm_averages}.

\subsection{Benchmark details\label{sec:benchmark}}

\subsubsection{Summary list}
We briefly describe in Table \ref{tab:tasks_summary} the different combinatorics task addressed in this work. These problems are related to known Erdős problems, although their exact formulations are adapted to the specific settings considered here and do not necessarily cover the most general cases.
\begin{table}[b]
\centering
\caption{Summary of mathematical problems used to evaluate CCTS. Each task defines a structured algorithm-discovery problem with a verifiable objective and a black-box evaluator.}
\small
\begin{tabular}{p{4.6cm} p{11.6cm}}
\hline
\textbf{Problem} & \textbf{Description} \\
\hline

Circle packing \cite{novikov2025alphaevolve} & Put $n=26$ circles inside of a unit square such that the sum of radii is maximized. Circles can be of different sizes but cannot overlap. \\

\href{https://www.erdosproblems.com/1097}{Arithmetic Kakeya Conjecture}
& Let $A$ be a set of $n$ integers. How many distinct $d$ can occur as the common difference of a three-term arithmetic progression in $A$? Are there always $O(n^{3/2})$ many such $d$? We consider an information-theoretic analogue of this question, replacing set cardinalities by Shannon entropies of linear projections, in the spirit of the entropy sumset theory developed in \cite{tao2010sumset}.\\


\href{https://www.erdosproblems.com/507}{Heilbronn's Triangle Problem}
& Places 11 points inside a unit-area equilateral triangle to maximize the minimum area of any triangle formed by three points. Out-of-bounds points are projected back, and the score is the smallest triangle area. \\

\href{https://www.erdosproblems.com/106}{Squares-in-Square}
& Places $n$ squares with disjoint interiors inside a unit square to maximize the sum of side lengths. Overlapping interiors are forbidden; touching boundaries is allowed. \\

\hline
\end{tabular}
\label{tab:tasks_summary}
\end{table}

\subsubsection{Circle Packing\label{sec:circle_packing}}

Here the goal is to pack \(26\) circles inside the unit square \([0,1]^2\). Each circle is specified by a center \(c_i=(x_i,y_i)\in\mathbb{R}^2\) and a radius \(r_i\ge 0\). A packing is feasible if every circle lies entirely in \([0,1]^2\) and circles do not overlap (touching is allowed). The optimization objective is to maximize the total size (total radius mass) of the packing which can be stated as follows.
\begin{align}
\sum_{i=1}^{26} r_i .
\end{align}

\begin{figure}[h]
    \centering
    \includegraphics[width=0.7\textwidth]{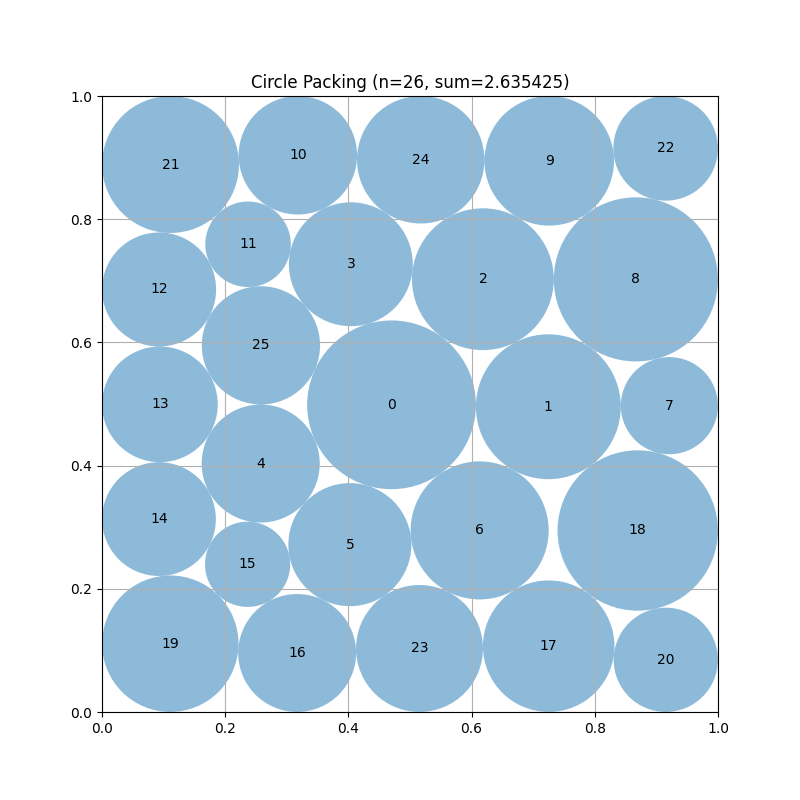}
    \caption{Optimal circle packing found by CCTS after 120 iterations, consistently with what has been found with AlphaEvolve \cite{novikov2025alphaevolve}.}
    \label{fig:circle_packing_results}
\end{figure}

This serves as a simple scalar measure of how much circular area can be accommodated in the square under these constraints. We report in Fig.~\ref{fig:circle_packing_results} the solution found for the circle packing task, which reproduces the optimal result reported in AlphaEvolve \cite{novikov2025alphaevolve}. 

\subsubsection{Arithmetic Kakeya Conjecture}

This problem concerns a discrete random vector $(X,Y)$ with finite support and asks how large the entropy of one linear projection (e.g.\ a difference) can be relative to the entropies of several other linear projections (e.g.\ coordinates and sums).
Formally, for linear maps $\pi_r(X,Y)=X+rY$ (with $\pi_\infty(X,Y)=Y$), one seeks sharp constants $C$ (sum-difference constant \cite{kakeya_tao2025sum}) such that,
\begin{align}
H(\pi_{-1}(X,Y))
\;\le\;
C \max\{H(\pi_0(X,Y)),\,H(\pi_1(X,Y)),\,H(\pi_2(X,Y)),\,H(\pi_\infty(X,Y))\} ,
\end{align}

holds for all finitely supported discrete $(X,Y)$ \cite{georgiev2025mathematical, kakeya_tao2025sum}. This entropy inequality is an information-theoretic analogue of classical sum-difference and arithmetic progression problems \cite{tao2010sumset}, including Erd\H{o}s Problem~\#1097  \cite{ai-erdos-2026}.

Here a candidate solution specifies a finitely supported probability distribution on $\mathbb{Z}^2$, interpreted as the joint law of $(X,Y)$. The evaluator computes the Shannon entropies of the projected random variables,

\begin{align}
X-Y,\quad X,\quad X+Y,\quad X+2Y,\quad Y ,
\end{align}
by aggregating probability mass along each projection \cite{kakeya_tao2025sum, kakeya_green2019arithmetic}. The score assigned to a candidate is the entropy ratio,

\begin{align}
\frac{H(X-Y)}{\max\{H(X),\,H(X+Y),\,H(X+2Y),\,H(Y)\}} .
\end{align}

Maximizing this ratio corresponds to constructing distributions where the difference projection is unusually complex compared to the other directions. Because the theoretical constant is defined as a supremum over all discrete distributions, each candidate’s entropy ratio provides a valid lower bound on the corresponding sum-difference constant.

\subsubsection{Heilbronn's Triangle Problem}

The Heilbronn triangle problem asks for a placement of $11$ points inside a fixed equilateral triangle of area $1$ that makes every triangle formed by three of the points ``as large as possible'' in the worst case \cite{georgiev2025mathematical}. Hence fixing the unit-area equilateral triangle $K\subset\mathbb{R}^2$, the task is to choose,
\begin{align}
P=\{p_1,\dots,p_{11}\}\subset K ,
\end{align}
so as to maximize the smallest area of any triangle formed by three selected points,
\begin{align}
m(P)\;=\;\min_{1\le i<j<k\le 11}\ \mathrm{Area}(p_i,p_j,p_k),
\qquad
\text{maximize } m(P).
\end{align}
Intuitively, good configurations avoid having any three points nearly collinear or overly clustered, since that would create a very small triangle and reduce $m(P)$.

In the testing a candidate returns an array of $11$ planar points (shape $(11,2)$). Following the description mentioned in \cite{georgiev2025mathematical} the evaluator then,
\begin{itemize}
  \item \textbf{Feasibility check:} if the output does not have the correct shape or contains non-finite entries, it is marked invalid (score $0$).
  \item \textbf{Repair step:} if any point lies outside $K$, it is projected to the nearest point on the boundary of $K$ (producing a repaired set $\tilde P$).
  \item \textbf{Score computation:} the score is the minimum area over all $\binom{11}{3}$ triangles formed by triples of repaired points \cite{HeilbronnTriangleProblem_cohen2023new}:
  \begin{align}
    \text{score} \;=\; m(\tilde P)
    \;=\;\min_{i<j<k}\ \mathrm{Area}(\tilde p_i,\tilde p_j,\tilde p_k),
  \end{align}
  where (in 2D) $\mathrm{Area}(a,b,c)=\tfrac12\bigl|\,(b-a)\times(c-a)\,\bigr|$.
\end{itemize}
Note that in this case higher scores are better.

\subsubsection{Squares-in-Square}

This Erd\H{o}s Problem asks how large the total side length of \(n\) squares can be when all squares are placed inside the unit square \([0,1]^2\) with disjoint interiors \cite{georgiev2025mathematical}. Disjoint interiors means that squares may touch along edges or at points, but any overlap of positive area is forbidden. Let \(C(n)\) denote the maximum achievable sum of side lengths under these constraints. Erd\H{o}s specifically asked whether, for \(n = k^2 + 1\), one always has \(C(n) = k\), extending the obvious construction showing \(C(k^2) = k\) \cite{ErdosProblemsWebsite, georgiev2025mathematical}.

Here a candidate outputs \(n\) squares, each represented by a tuple \((c_x,c_y,\theta,s)\) specifying its center, rotation angle, and side length. Following the description stated in \cite{georgiev2025mathematical}, the evaluator first checks validity: all square vertices must lie within \([0,1]^2\), side lengths must be nonnegative, and no two squares may intersect with positive area (boundary contact is allowed). If any of these conditions fails, the configuration is declared invalid and assigned score \(-\infty\). If the configuration is valid, the score is simply the sum of side lengths,
\begin{align}
\text{score} \;=\; \sum_{i=1}^n s_i.
\end{align}

Thus, higher scores correspond to packings that place more total square edge length into the unit square without violating the constraints. The basic formulation for the score is the same as that described in \cite{georgiev2025mathematical}. Within this formulation, in addition to checking the score, it is necessary to verify that the produced solution is optimal.



\clearpage

\end{document}